\documentclass[preprint,12pt]{elsarticle}
\usepackage{booktabs,tabularx,rotating,url}
\usepackage{longtable}
\usepackage{graphicx}
\usepackage[labelfont=bf,labelsep=period]{caption}

\usepackage{array}
\usepackage{caption}
\usepackage{amssymb}
\usepackage{amsmath}

\journal{Biomedical Signal Processing and Control}

\begin{document}

\begin{frontmatter}



\title{Exploring the Challenge and Value of Deep Learning in Automated Skin Disease Diagnosis}


\author[label1]{Runhao Liu\corref{cor1}\fnref{cofirst}}
\author[label2]{Ziming Chen\fnref{cofirst}}
\author[label3]{Guangzhen Yao\fnref{cofirst}}
\author[label4]{Peng Zhang\corref{cor1}}

\fntext[cofirst]{These authors contributed equally.}

\cortext[cor1]{Corresponding author.\\
\hspace*{4ex}\textit{Email addresses}: runhaoliu@zju.edu.cn (R. Liu), pengz@zju.edu.cn (P. Zhang)}

\affiliation[label1]{organization={Polytechnic Institute, Zhejiang University},
            city={Hangzhou},
            postcode={310015}, 
            country={China}}
\affiliation[label2]{organization={School of Social and Public Administration, East China University Of Science And Technology},
            city={Shanghai},
            postcode={200237}, 
            country={China}}
\affiliation[label3]{organization={College of Science, National University of Defense Technology},
            city={Changsha},
            postcode={410073}, 
            country={China}}
\affiliation[label4]{organization={School of Mathematical Sciences, Zhejiang University},
            city={Hangzhou},
            postcode={310058}, 
            country={China}}




\begin{abstract}
Skin cancer is one of the most prevalent and deadly forms of cancer worldwide, highlighting the critical importance of early detection and diagnosis in improving patient outcomes. Deep learning (DL) has shown significant promise in enhancing the accuracy and efficiency of automated skin disease diagnosis, particularly in detecting and classifying skin lesions. However, several challenges remain for DL-based skin cancer diagnosis, including complex features, image noise, intra-class variation, inter-class similarity, and data imbalance. This review synthesizes recent research and discusses innovative approaches to address these challenges, such as data augmentation, hybrid models, and feature fusion. Furthermore, the review highlights the integration of DL models into clinical workflows, offering insights into the potential of deep learning to revolutionize skin disease diagnosis and improve clinical decision-making. This review uniquely integrates a PRISMA-based methodology with a challenge-oriented taxonomy, providing a systematic and transparent synthesis of recent deep learning advances for skin disease diagnosis. It further highlights emerging directions such as hybrid CNN-Transformer architectures and uncertainty-aware models, emphasizing its contribution to future dermatological AI research.
\end{abstract}

\begin{graphicalabstract}
\end{graphicalabstract}

\begin{highlights}
\item Systematically reviews 91 studies (2020–2025.6), identifying four core challenges in deep learning-based skin cancer diagnosis: complex features, image noise, intra-class similarity, and class imbalance.
\item Summarizes advanced solutions including data augmentation (GANs, diffusion models), hybrid CNN-Transformer architectures, and multimodal fusion for robust and generalizable diagnostic performance.
\item Highlights optimized loss functions, transfer learning, and attention mechanisms as critical strategies to improve class balance, interpretability, and clinical applicability in automated dermatology.
\end{highlights}


\end{frontmatter}



\section{Introduction}
\label{sec1}
As a major global health burden, skin cancer has seen a steady rise in incidence rates in recent years \cite{Pathan}. Based on surveys, nearly 108,420 new cases of skin cancer were diagnosed in the United States in 2020 \cite{R. L. Siegel}, while there are about 97,610 new cases confirmed, with 7,990 deaths estimated in 2022 \cite{M. L. Hedberg}. Skin cancer is a rapidly growing disease characterized by the abnormal and uncontrolled proliferation of skin cells, primarily caused by prolonged exposure to ultraviolet (UV) radiation from the sun. If left untreated, these cancerous cells may invade deeper tissues and spread to other parts of the body, leading to severe complications. This disease is broadly categorized into two main types: melanoma and non-melanoma skin cancer. Melanoma is the most aggressive and life-threatening form of skin cancer. It often appears as a dark, irregularly shaped spot on the skin, which may develop from a pre-existing mole or arise as a new lesion. Non-melanoma skin cancers are more common but generally less aggressive than melanoma. These cancers typically present as persistent lesions, scaly patches, or nodules on sun-exposed areas of the skin \cite{M. Q. Khan}. In addition, skin cancer is also typically categorized into two distinct forms: benign and malignant. As a lethal malignancy, skin cancer mainly encompasses 7 types: actinic keratoses (AKIEC), basal cell carcinoma (BCC), benign keratosis-like lesion (BKL), dermatofibroma (DF), melanoma (MEL), melanocytic nevi (NV), and vascular lesion (VASC), all of which can cause serious health risks \cite{U. Leiter}.

Early diagnosis of skin cancer plays a critical role in reducing mortality rates and improving patient outcomes. When skin cancer is identified at an early stage, it can typically be treated effectively through essential excision, a relatively simple surgical procedure with high success rates \cite{K. G. Lewis}. Early detection not only minimizes the need for aggressive interventions but also significantly lowers the risk of metastasis, thereby improving prognosis and overall survival rates. Conversely, delayed diagnosis of skin cancer, particularly malignant melanoma, is associated with a dramatically increased risk of mortality. Patients diagnosed at an advanced stage face an over 80$\%$ likelihood of death due to the aggressive nature of metastatic melanoma. Once the cancer has penetrated deeper layers of the skin and spread to lymph nodes or distant organs, treatment options become significantly more complex and less effective. Late-stage management often necessitates extensive surgical excision, systemic chemotherapy, immunotherapy, or radiation therapy, all of which can lead to substantial physical and financial burdens \cite{C. M. Balch}. Therefore, early recognition and diagnosis of skin cancer are paramount in mitigating disease progression, reducing mortality risk, and alleviating the social and economic impact of advanced cancer care. Comprehensive efforts from individuals, healthcare professionals, and public health organizations are essential in facilitating early intervention and improving overall patient outcomes.

As a primary technology for the diagnosis of skin cancer, dermoscopy is commonly utilized. However, interpreting dermoscopic images relies heavily on the expertise of professional dermatologists. Although diagnostic accuracy can reach 75\%--84\% in expert hands, manual assessment is hindered by subjectivity, heavy reliance on experience, and limited reproducibility \cite{G. Argenziano}, leading to variability in diagnostic efficiency. Nevertheless, while CAD has the potential to enhance accessibility and accuracy in skin cancer diagnosis \cite{A. Maiti}, numerous technical challenges remain that hinder its widespread effectiveness and reliability. First, uneven lesion color and blurred edges make segmentation and classification difficult. Intra-class variation and inter-class similarity further complicate differentiation, as lesions of the same type can look different, while benign and malignant ones may appear similar. Additionally, image noise and artifacts, such as hair, skin texture, and lighting variations, interfere with accurate detection. The diversity of lesion appearances across different skin tones and demographics adds another layer of complexity. Moreover, the limited generalizability of traditional CAD models, which rely on hand-crafted features and rule-based algorithms, reduces their adaptability to real-world clinical settings. These challenges highlight the need for more advanced deep learning models and improved image processing techniques to enhance CAD accuracy and reliability in skin disease diagnosis.

\begin{figure}[htbp]
    \centering
    \includegraphics[width=0.7\textwidth]{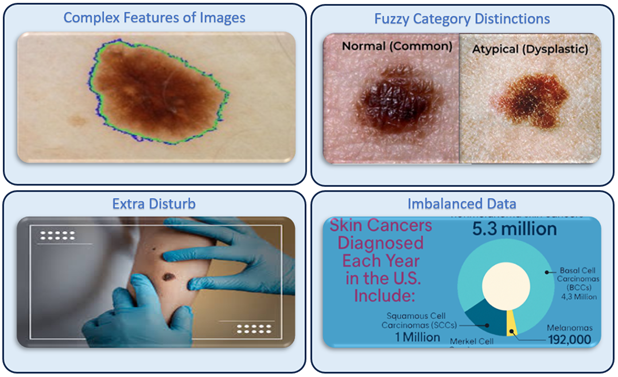}
    \caption{Challenges of current deep learning studies.}
    \label{challenges_of_curreng}
\end{figure}

With the development of artificial intelligence (AI), deep learning has significantly advanced the diagnosis of dermatological diseases by improving accuracy and efficiency in medical image analysis \cite{M. de Bruijne}. Instead of traditional methods, which rely on handcrafted features, deep learning models can automatically extract complex patterns from large-scale medical datasets, enabling more precise differentiation between benign and malignant skin conditions. Through iterative learning and iterative training, these models adapt to variations in lesion appearance across different patients, overcoming challenges such as uneven color, blurred edges, and visual similarities between different skin diseases \cite{A. Bouzid-Daho}. Moreover, deep learning provides higher diagnostic accuracy, sometimes surpassing human dermatologists in specific tasks \cite{J. Kaur}. By leveraging vast amounts of medical data, these models enhance early detection and classification, reducing the risk of misdiagnosis and enabling timely intervention. Another key advantage is the automation of image analysis, which reduces the workload on healthcare professionals and improves diagnostic efficiency, especially in areas with limited access to specialized dermatologists \cite{R. R. Kumar}. With the rapid expansion of medical data, deep learning plays an essential role in modernizing skin disease diagnosis, offering fast, intelligent, and scalable solutions that enhance clinical decision-making and patient outcomes \cite{J. S. H. Baxter}. However, as shown in Fig.\ref{challenges_of_curreng}, besides the common challenges required to be solved by computer-aided diagnosis, deep learning is facing other obstacles, imbalanced data in particular, which causes a serious overfit problem. Thus, ongoing research is necessary to further refine these models and address challenges such as image noise and variability in skin lesion presentations.

In contrast to previous reviews, this study not only categorizes recent works under four fundamental challenges but also incorporates a PRISMA-driven selection process to ensure methodological transparency and reproducibility. This dual framework provides both a conceptual and procedural contribution to dermatological AI research.

\subsection{Objectives}
\label{Objectives}
This review aims to analyze the advancements and challenges in the application of deep learning to automated skin disease diagnosis. By evaluating the latest research from 2020 to 2025, the review aims to highlight the effectiveness of various deep learning models in enhancing diagnostic accuracy, overcoming image processing challenges, and addressing data imbalance. Additionally, the review seeks to explore innovative methodologies, including data augmentation, hybrid models, and optimization techniques, that contribute to improving the performance and clinical applicability of deep learning in dermatological diagnostics.
\subsection{Contribution}
\label{Contribution}
We identified the persisting challenges and limitations in the application of deep learning models for skin disease diagnosis, including complex features, image noise, intra-class variation and inter-class similarity, and data imbalance.

This review provides a comprehensive overview of the current research on how deep learning models address the four challenges mentioned above in automated skin disease diagnosis, synthesizing findings from various sources to offer a holistic perspective on the field. Then, we conduct a comparative analysis of various deep learning methodologies, offering insights into their strengths and weaknesses to guide researchers and clinicians in selecting the most appropriate approach.

Then, a comparative analysis of various deep learning methodologies, is conducted in this paper, offering insights into their strengths and weaknesses to guide researchers and clinicians in selecting the most appropriate approach. Compared with existing surveys, none of them integrates a PRISMA-based methodological rigor with a challenge-oriented taxonomy, nor do they systematically connect technical advances with clinical applicability.

With the current progress and challenges summarized, this review not only contributes to academic knowledge but also offers valuable guidance for future research directions and the clinical implementation of deep learning models in skin disease diagnosis.

The remainder of this paper is organized as follows.
Section~\ref{Methods} details the PRISMA-based methodology and literature search strategy.
Section~\ref{Findings} presents a bibliometric analysis of the selected studies, including publication trends and geographical distribution.
Section~\ref{Research data} introduces the benchmark datasets and common preprocessing techniques.
Section~\ref{Techniques for skin cancer diagnosis} provides a comprehensive review of deep learning methods structured around four key diagnostic challenges: complex features, image noise, intra-class variation, and class imbalance.
Section~\ref{Prospective techniques} analyzes the dominant technical strategies employed to address these challenges, such as data augmentation and hybrid models.
Section~\ref{Frontier Advances and Emerging Paradigms} discusses frontier advances and emerging paradigms, including multimodal fusion and uncertainty-aware learning.
Finally, Section~\ref{Discussion and Conclusion} offers a critical discussion on the historical progression and limitations, followed by the conclusion.

\section{Methodology}
\label{Methods}
\begin{figure}[htbp]
\centering
\includegraphics[width=1\textwidth]{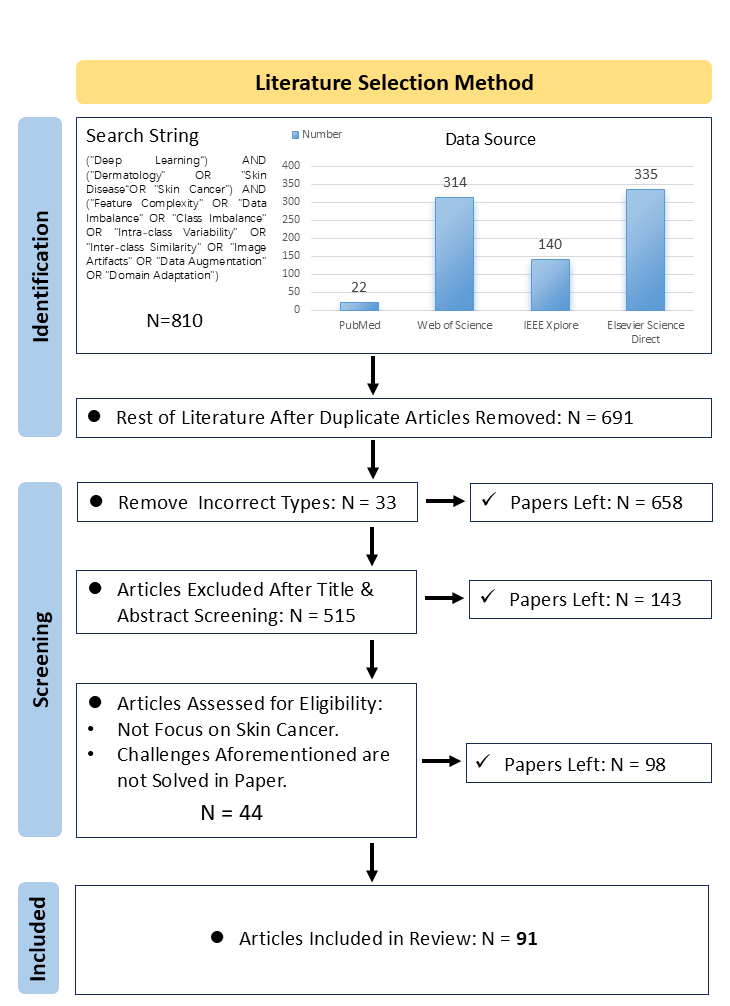}
\caption{The PRISMA flow of this literature.}\label{prisma}
\end{figure}

\subsection{PRISMA framework}
To conduct a more comprehensive review, the Preferred Reporting Items for Systematic Reviews and Meta-Analyses (PRISMA) framework was used for the literature retrieval and selection, as shown in Fig.\ref{prisma}. We performed 3 stages for this review. In the identification stage, this study searched the relevant literature through the databases of PubMed, Web of Science, IEEE Xplore, and Elsevier Science Direct, and initially obtained 810 records. Among them, Web of Science and Elsevier Science Direct provided more literature, contributing 314 and 335 articles, respectively, while PubMed and IEEE Xplore contributed 22 and 140 articles, respectively. After removing duplicates, a total of 691 articles were advanced to the next stage. A step-by-step screening strategy was utilized in the screening stage. First, 33 records of ineligible types were excluded, such as reviews, books, patents, conference abstracts, or papers without empirical data. This resulted in 658 documents. Subsequently, 515 irrelevant studies were excluded through title and abstract screening, as they either did not focus on deep learning or did not address dermatological diagnosis, leaving 143 articles for full-text assessment. In the final stage of eligibility assessment, we further excluded 44 papers with two main problems: first, the studies did not address dermatological diseases, and second, although the studies used deep learning, they did not address the core challenges, such as data imbalance and feature complexity. Initially, 98 studies met the inclusion criteria. Upon a secondary rigorous quality check, 7 additional studies were excluded due to insufficient data or lack of relevance, resulting in a final set of 91 articles included in this review. Zotero is utilized for literature collation and analysis. This structured and reproducible screening process minimizes selection bias and ensures that the included studies form a transparent and methodologically consistent evidence base for the subsequent analysis.

Instead of the usage of standard PRISMA screening procedure, the studies included in this review are organized according to four fundamental challenges frequently encountered in automated dermatological diagnosis, including complex features, image noise, intra and inter-class similarity, and data imbalance. This challenge-oriented structure allows the literature to be examined not merely by model type or algorithmic category like conventional surveys, but by the specific diagnostic obstacles each study seeks to address. Such a mapping facilitates a clearer comparison across heterogeneous methods and highlights how different technical strategies converge on common clinical difficulties. Consequently, this approach facilitates a more coherent synthesis of evidence throughout the review.

\subsection{Search strategy}
A comprehensive literature search was performed across multiple databases, including PubMed, Web of Science, IEEE Xplore, and Elsevier ScienceDirect. The search was limited to studies published between January 2020 and June 2025 to focus on the most recent advancements in deep learning for automated skin disease diagnosis.

The search utilized the following combination of keywords to ensure a comprehensive retrieval of studies. 

("Deep Learning") AND ("Dermatology" OR "Skin Disease" OR "Skin Cancer") AND ("Feature Complexity" OR "Data Imbalance" OR "Class Imbalance" OR "Intra-class Variability" OR "Inter-class Similarity" OR "Image Artifacts" OR "Data Augmentation" OR "Domain Adaptation")

This keyword strategy was specifically crafted to address key challenges in deep learning-based skin disease diagnosis, including feature complexity, data imbalance, intra-class variability, inter-class similarity, image artifacts, and domain adaptation. The inclusion of terms related to data augmentation also ensured that studies exploring methods to improve model performance were captured.

\subsection{Risk of Bias Assessment}
To assess the risk of bias in the included studies, we used Newcastle-Ottawa Scale (NOS). Moreover, risk of bias was assessed independently by two reviewers, who evaluated the studies in the following domains: selection bias, performance bias, detection bias, and reporting bias. Any disagreements between reviewers were resolved through discussion, and a third reviewer was consulted when consensus could not be reached.

\section{Findings}
\label{Findings}
\begin{figure}[htbp]
\centering
\includegraphics[width=0.7\textwidth]{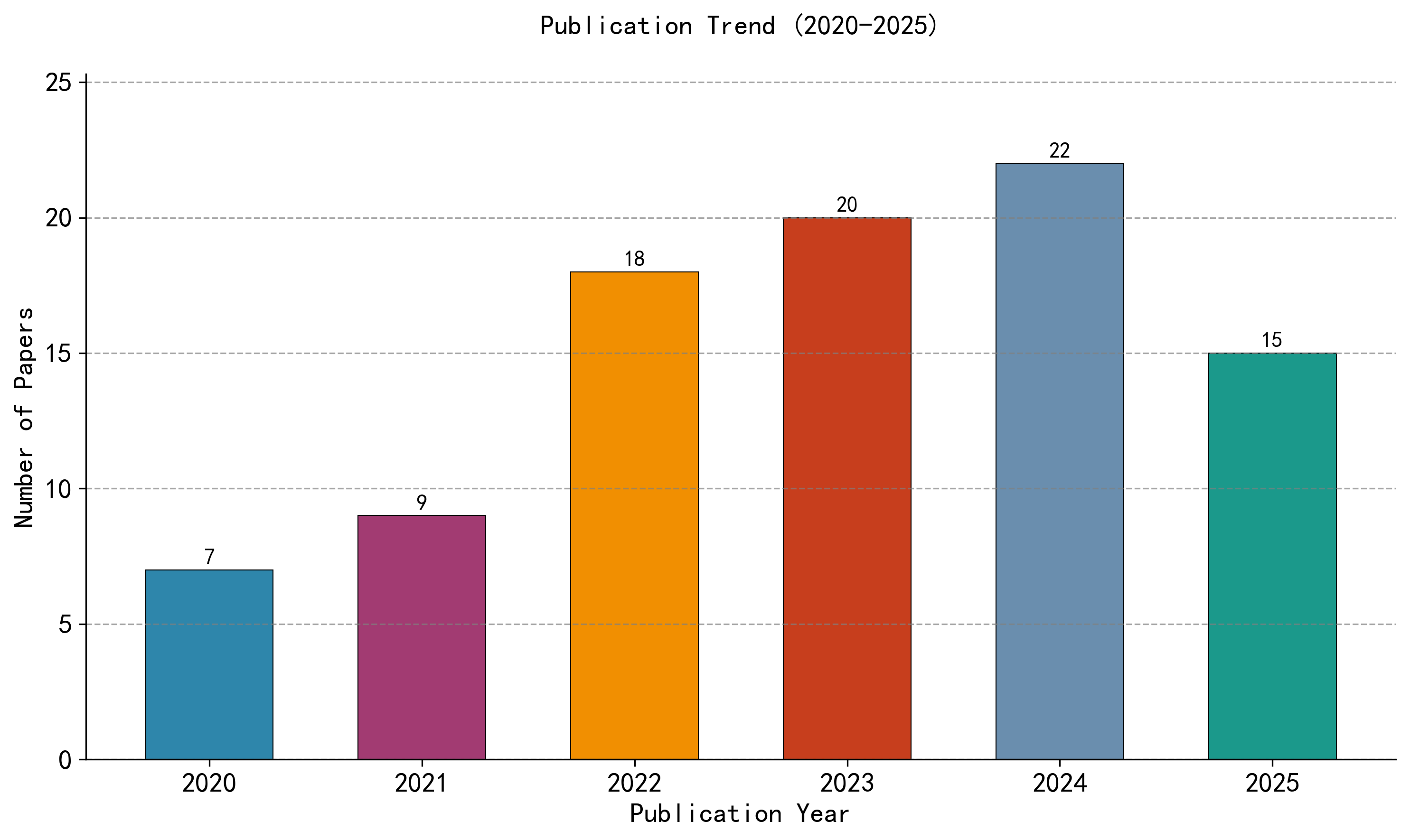}
\caption{Publication Trend of research from 2020 to June 2025.}\label{trend}
\end{figure}

We retrieved publications published from January 2020 to June 2025. As illustrated in Fig.~\ref{trend} that the publication trend indicates a significant and sustained increase in research activity concerning the application of deep learning to skin disease diagnosis. While the number of publications was relatively modest in 2020 ($n = 7$) and 2021 ($n = 9$), a marked surge occurred in 2022 with 18 papers published, followed by sustained high outputs in 2023 ($n = 20$) and 2024 ($n = 22$), suggesting the field entered a phase of accelerated development. This rapid growth may be attributed to advances in deep learning architectures and the availability 
of high-quality dermoscopic datasets. Although only 15 papers were published in the first half of 2025, this partial data is not indicative of a declining trend but rather reflects the incomplete nature of the annual publication cycle. Overall, the trend reflects increasing academic attention and clinical relevance, with a shift from foundational model development towards addressing key diagnostic challenges, such as feature complexity and data imbalance, and facilitating the translation of AI models into clinical practice.

\begin{figure}[htbp]
\centering
\includegraphics[width=0.8\textwidth]{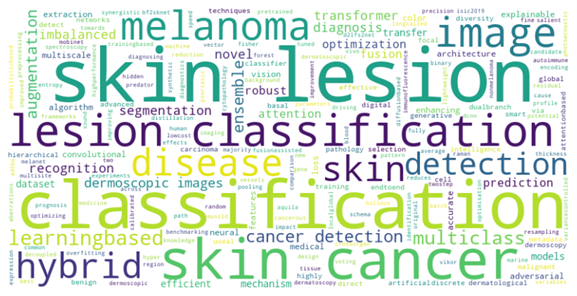}
\caption{Word Cloud of the reviewed studies.}\label{word_cloud}
\end{figure}

To visualize the thematic focus of the included studies, Fig.\ref{word_cloud} presents a word cloud generated from the article titles. The font size is proportional to the frequency of the terms. The high frequency of the terms “skin”, “cancer”, “disease”, and “lesion” indicates that the research mainly focuses on the automatic detection and diagnosis of skin lesions, especially skin cancer. The prominence of keywords such as “classification”, “detection”, and “segmentation” indicates that the research focuses on the classification, detection, and image segmentation of skin lesions to improve diagnostic accuracy. In addition, the words “learningbased”, “hybrid”, “ensemble”, and “transformer” indicate that researchers are exploring different deep learning architectures as well as hybrid models to optimize the diagnostic performance. Furthermore, terms such as “imbalance”, “augmentation”, and “explainable” indicate that research is still facing the challenges of data imbalance, feature complexity, and model explainability, and some research attempts to improve the robustness and clinical usability of the models through data augmentation, migration learning, and federated learning. Overall, the research direction in this field covers disease classification, detection optimization, deep learning method improvement, and is constantly exploring stronger models and better data processing solutions to promote the application of intelligent medical image analysis in dermatological diagnosis.

The selected studies were predominantly published in high-impact JCR Q1 journals, as listed in Table \ref{tab:top10journals}. Among which, IEEE ACCESS has published 7 studies, which indicate the wide application of computer vision and multimedia processing technology in dermatological diagnosis. In addition, journals such as MULTIMEDIA TOOLS AND APPLICATIONS, BIOMEDICAL SIGNAL PROCESSING AND CONTROL, and SCIENTIFIC REPORTS occupy an essential position in the fields of computer-aided diagnosis (CAD), medical image analysis, and clinical medical application, further reflecting the interdisciplinary integration trend in this field. The trend is toward interdisciplinary convergence in the field. The research involves a variety of disciplines such as computer science, engineering, biomedicine, and medicine, indicating that the application of deep learning in dermatologic diagnosis has gradually transformed from methodological research to clinical practice.

\begin{table}[htbp]
\centering
\caption{Top 10 Journals / Conferences in included articles.}
\label{tab:top10journals}
\begin{tabular}{
>{\centering\arraybackslash}m{6cm} 
>{\centering\arraybackslash}m{3cm} 
>{\centering\arraybackslash}m{2cm} 
>{\centering\arraybackslash}m{1.5cm}}
\toprule
\textbf{Journals / Conferences} & \textbf{Subject field} & \textbf{JCR Quartiles} & \textbf{Counts} \\
\midrule
IEEE ACCESS & Engineering & Q1 & 8 \\
BIOMEDICAL SIGNAL PROCESSING AND CONTROL & Biomedical & Q1 & 5 \\
MULTIMEDIA TOOLS AND APPLICATIONS & Engineering & Q1 & 4 \\
IEEE JOURNAL OF BIOMEDICAL AND HEALTH INFORMATICS & Engineering & Q1 & 3 \\
COMPUTERS IN BIOLOGY AND MEDICINE & Biology & Q1 & 3 \\
SCIENTIFIC REPORTS & Multidisciplinary & Q1 & 3 \\
DIAGNOSTICS & Medicine & Q2 & 3 \\
COMPUTER METHODS AND PROGRAMS IN BIOMEDICINE & Computer Science & Q1 & 3 \\
COMPUTERS IN BIOLOGY AND MEDICINE & Computer Science & Q1 & 3 \\
CMC-COMPUTERS MATERIALS \& CONTINUA & Mathematics & Q1 & 2 \\
\bottomrule
\end{tabular}
\end{table}

\begin{figure}[htbp]
\centering
\includegraphics[width=1\textwidth]{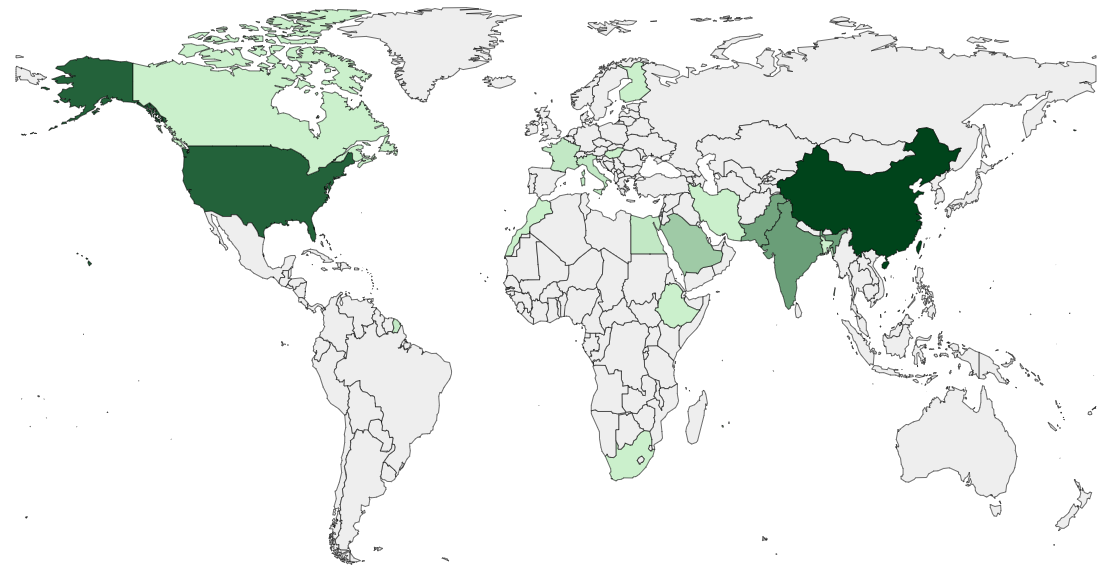}
\caption{Global distribution of first authors’ affiliations.}\label{map}
\end{figure}

Fig.\ref{map} illustrates the global distribution of the first authors’ affiliations across the 91 reviewed studies on deep learning-based automated skin disease diagnosis. The heatmap highlights that research contributions are predominantly concentrated in China with 24 papers and the United States, who possesses 20 publication, followed by significant participation from India, Canada, Saudi Arabia, and several European countries. Other regions, including parts of Africa and South America, show relatively lower research activity. This trend indicates that regions with advanced medical imaging infrastructure and robust AI research ecosystems dominate the field. In contrast, contributions from Africa and South America remain sparse, suggesting potential disparities in dermatological data availability and research funding.

\section{Research data}
\label{Research data}
High quality dataset is required in the study of skin disease diagnosis based on deep learning. Thus, we summarized benchmark datasets commonly utilized for experimentation, as shown in Table \ref{tab:benchmarkdatasets}. 

Table \ref{tab:benchmarkdatasets} provides an overview of widely used benchmark datasets in automated skin lesion detection research, including Skin Imaging Collaboration(ISIC) Series, HAM10000, PH2, Derm7pt, and SkinCon. These datasets serve as essential resources for developing and evaluating deep learning models in dermatological diagnosis. The ISIC Series, curated by the International Skin Imaging Collaboration, encompasses a diverse range of skin lesion types but is characterized by extreme class imbalance, which poses challenges for model training. Similarly, HAM10000, a large-scale dataset with 10015 dermoscopic images spanning seven common pigmented skin lesion types, also exhibits class imbalance, potentially leading to biased model predictions. In contrast, PH2 is a smaller dataset with 200 images, offering a well-balanced class distribution, making it particularly suitable for model validation and fairness assessment. Additionally, Derm7pt is distinguished by its seven-point checklist annotation system, providing 1011 dermoscopic images with both clinical and imaging data, making it a valuable resource for multimodal deep learning studies. The SkinCon dataset contains 3230 images spanning 48 clinical concepts, enabling fine-grained lesion classification and concept-level labeling. While these datasets significantly contribute to the advancement of deep learning in dermatology, they also present challenges such as class imbalance, small sample sizes, and the need for clinical applicability. Researchers should carefully select appropriate datasets based on their study objectives and employ techniques such as data augmentation, resampling strategies, and multimodal learning to enhance model robustness and improve diagnostic performance in clinical applications.

It is worthwhile to mention that among these datasets, ISIC is more widely used. The dataset is updated continuously. The imbalance characteristic is more obvious in the series compared to other datasets. For instance, the ISIC-2020 contains 32507 samples in total and 2 classes, where the number of benign and malignant images is 31926 and 581, respectively. Besides, the intra-class variability and inter-class similarity are also significant in this dataset. Without technologies to cope with these obstacles, models with low performance could be acquired \cite{V. Rotemberg}.


\begin{sidewaystable}[htbp]
\centering
\caption{Mainstream benchmark datasets.}
\label{tab:benchmarkdatasets}
\begin{tabular}{
>{\centering\arraybackslash}m{2.8cm}
>{\centering\arraybackslash}m{5.8cm}
>{\centering\arraybackslash}m{3.2cm}
>{\centering\arraybackslash}m{2.8cm}
>{\centering\arraybackslash}m{5.5cm}}
\toprule
\textbf{Dataset} & \textbf{Introduction} & \textbf{Characteristic} & \textbf{Segmentation Mask} & \textbf{Source} \\
\midrule

ISIC Series &
Representative example: ISIC-2020 contains \textbf{33,126 images} with \textbf{2 classes} (benign 31,926; malignant 581). Earlier years (2016–2017) include more classes and pixel-level lesion masks. &
Extremely imbalanced; multi-category across years. &
Available (ISIC-2016/2017);  
Not available in ISIC-2020. &
\url{https://challenge.isic-archive.com} \\
\midrule

HAM10000 &
Contains \textbf{10,015 images} across \textbf{7 classes} (AKIEC, BCC, BKL, DF, MEL, NV, VASC). Widely used for benchmark classification tasks. &
Obvious imbalance; seven diagnostic categories. &
Not provided. &
\url{https://www.kaggle.com/kmader/skin-cancer-mnist-ham10000} \\
\midrule

PH2 &
Includes \textbf{200 dermoscopic images} (common nevi + melanomas) with high-quality lesion annotation. &
Small dataset; balanced categories. &
Provided (pixel-level). &
\url{https://www.fc.up.pt/addi/ph2} \\
\midrule

Derm7pt &
Contains \textbf{1,011 images} labeled according to the \textbf{7-point checklist}, includes clinical metadata. &
Multimodal annotation; diagnostic criteria aligned with clinical practice. &
Not provided. &
\url{https://derm7pt.org} \\
\midrule

SkinCon &
Includes \textbf{3,230 clinical images} covering \textbf{48 dermatological concepts}, supporting concept-level modeling. &
Concept-level labeling; multiple clinical attributes. &
Not provided. &
\url{https://skincon-dataset.com} \\
\bottomrule
\end{tabular}
\end{sidewaystable}

\begin{figure}[htbp]
\centering
\includegraphics[width=1\textwidth]{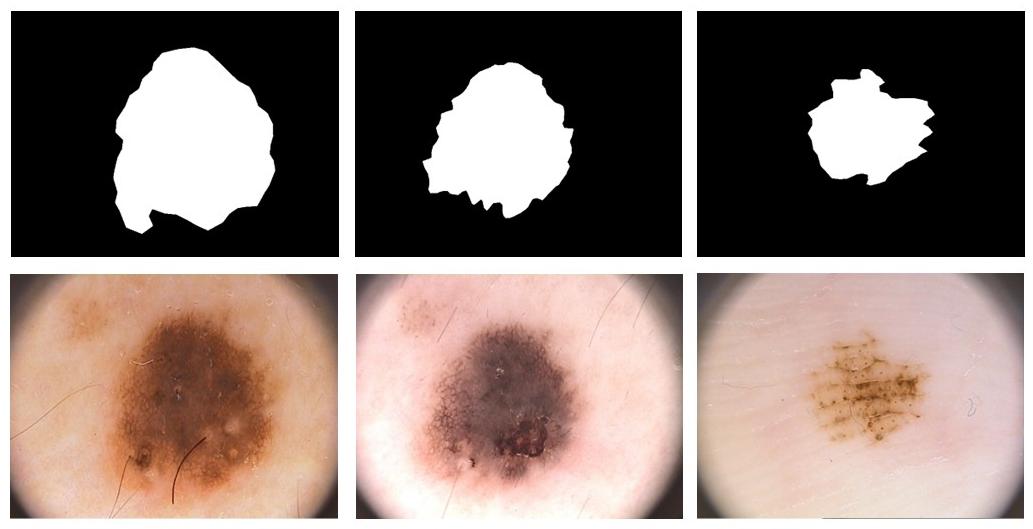}
\caption{Representative samples from major benchmark skin lesion datasets.}\label{data_demonstration_fig}
\end{figure}

Representative examples of dermoscopic images and their corresponding segmentation masks are shown in Fig.\ref{data_demonstration_fig}. The upper row presents the annotated lesion regions, and the lower row shows the original images. These samples illustrate typical variations in lesion color, texture, and boundary clarity, reflecting the visual diversity found in common benchmark datasets.

The analysis of benchmark datasets for deep learning-based skin disease diagnosis shows that data quality and balance are crucial for model performance. Large datasets like ISIC Series and HAM10000 are widely used but suffer from severe class imbalance, which can lead to biased predictions. Smaller datasets like PH2 are more balanced and useful for validation. Derm7pt provides multimodal data, supporting more detailed diagnosis, while SkinCon allows for fine-grained classification. ISIC datasets also pose challenges with intra-class variability and inter-class similarity, requiring data augmentation and advanced techniques to improve model performance. Choosing the right dataset and preprocessing methods is essential for building reliable and clinically useful deep learning models.

\section{Techniques for skin cancer diagnosis}
\label{Techniques for skin cancer diagnosis}
\subsection{Traditional method}
\label{Traditional method}
Traditional dermatological diagnosis primarily relies on dermoscopy, clinical examination, and histopathological analysis, combined with the subjective judgment of medical professionals. This process demands a high level of expertise and experience from dermatologists, making it time-consuming and prone to inter-observer variability \cite{V. Singh}. Dermoscopy, a non-invasive imaging technique, enhances the visibility of subcutaneous structures by reducing surface reflection, thereby improving the detection of lesion details. Dermatologists employ several diagnostic frameworks to interpret dermoscopic images. The ABCD Rule is most widely used in a common clinical environment. This method assesses asymmetry, border irregularity, color variation, and lesion diameter to estimate malignancy risk \cite{F. Nachbar}. Menzies' method focuses on the evolutionary changes of the lesion, such as pigment distribution and structural transformation over time \cite{S. W. Menzies}. Besides, the Seven-Point Checklist is a structured approach evaluating seven key dermoscopic criteria, including atypical pigment patterns, disappearance of pigment networks, and vascular irregularities \cite{G. Argenziano}. Despite the clinical effectiveness of these methods, they are highly dependent on the skill and experience of the clinician, leading to potential diagnostic inconsistencies and errors. Moreover, challenges such as uneven pigmentation, blurred lesion borders, and external obstructions such as hair, lighting conditions, can further complicate manual diagnosis.

\begin{figure}[htbp]
\centering
\includegraphics[width=1\textwidth]{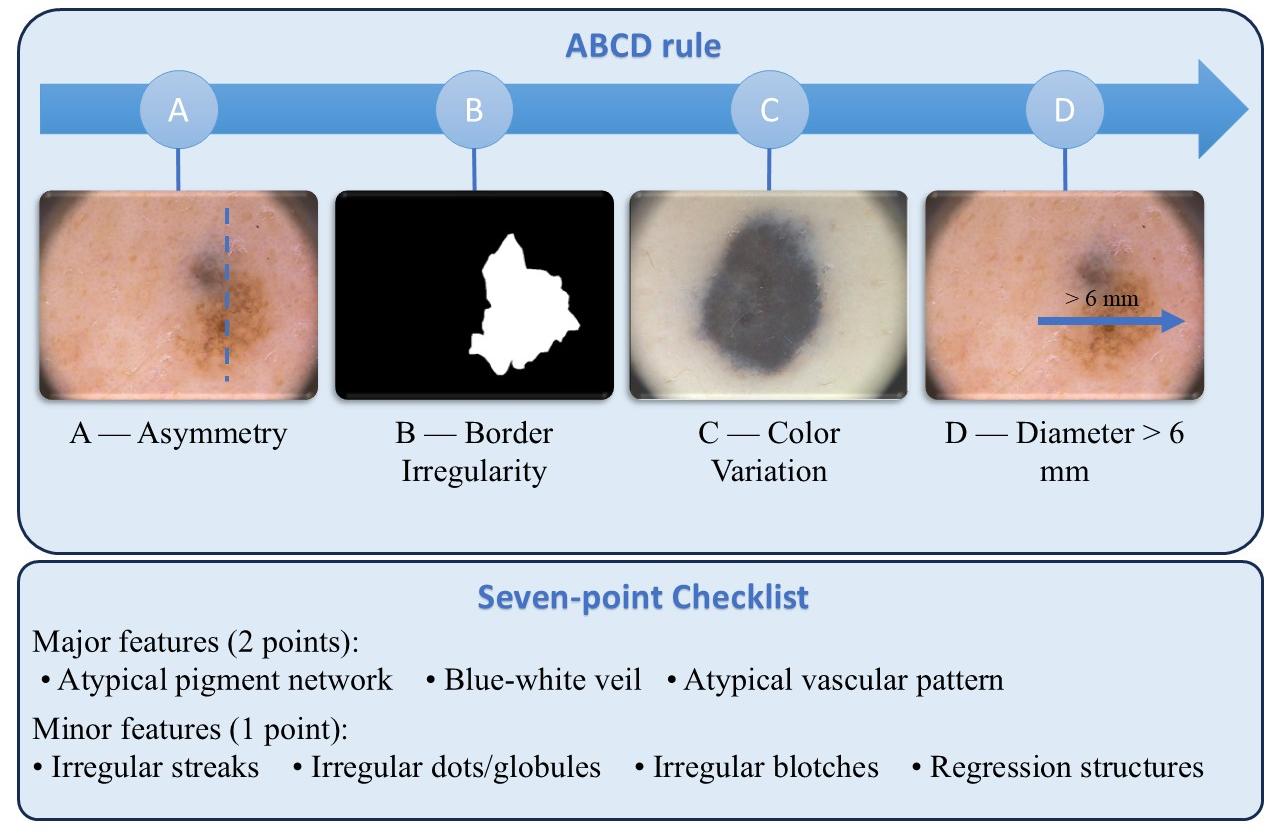}
\caption{Visual illustration of common dermoscopic diagnostic criteria.}\label{common_dermoscopic_diagnostic_criteria}
\end{figure}

A schematic illustration of the dermoscopic diagnostic criteria is presented in Fig.\ref{common_dermoscopic_diagnostic_criteria}. The upper panel summarizes the ABCD rule with representative examples, and the lower panel outlines the major and minor features of the seven-point checklist. These visual elements help clarify the traditional clinical assessment methods used before deep learning–based analysis.

\subsection{Early CAD method}
\label{Early CAD method}
With advancements in computer science, CAD systems emerged to overcome the limitations of manual assessment. The early development of CAD for dermatological diseases primarily focused on traditional image segmentation and feature extraction methods to analyze dermoscopic images. Initially, threshold segmentation, region growing, and edge detection techniques were widely used to identify melanoma lesions. However, these approaches relied heavily on manually extracted features, such as skin color and texture, leading to complex implementation processes, high computational costs, and poor robustness when dealing with variations in lesion appearance, uneven pigmentation, or external occlusions In the early stages, CAD systems aimed to classify dermoscopic images based on handcrafted features, improving upon traditional diagnostic methods such as the ABCD rule, Menzies method, and the seven-point checklist. Several researchers proposed advancements in this domain. Lazebnik et al. introduced a technique using approximate global geometric correspondence, segmenting input images into fine-grained subregions and computing local feature histograms, which enhanced classification accuracy \cite{S. Lazebnik}. Ballerini et al. developed more precise and objective feature extraction methods, reducing reliance on human-derived lesion characteristics. Their work improved the performance of the k-nearest neighbor (KNN) algorithm in dermoscopic image classification, addressing limitations such as subjectivity, inaccuracies, and weak robustness in manual feature extraction \cite{L. Ballerini}. Barata et al. proposed a bag-of-features approach based on local image descriptors, using color and texture comparisons to evaluate their discriminative power in lesion classification \cite{C. Barata}. Vasconcelos et al. enhanced CAD-based melanoma diagnosis by assessing lesion color variations according to the ABCD rule, demonstrating that color feature evaluation was highly effective for non-invasive dermoscopic assessment \cite{M. J. M. Vasconcelos}.

\subsection{Deep learning method}
\label{Deep learning method}
\begin{figure}[htbp]
\centering
\includegraphics[width=1\textwidth]{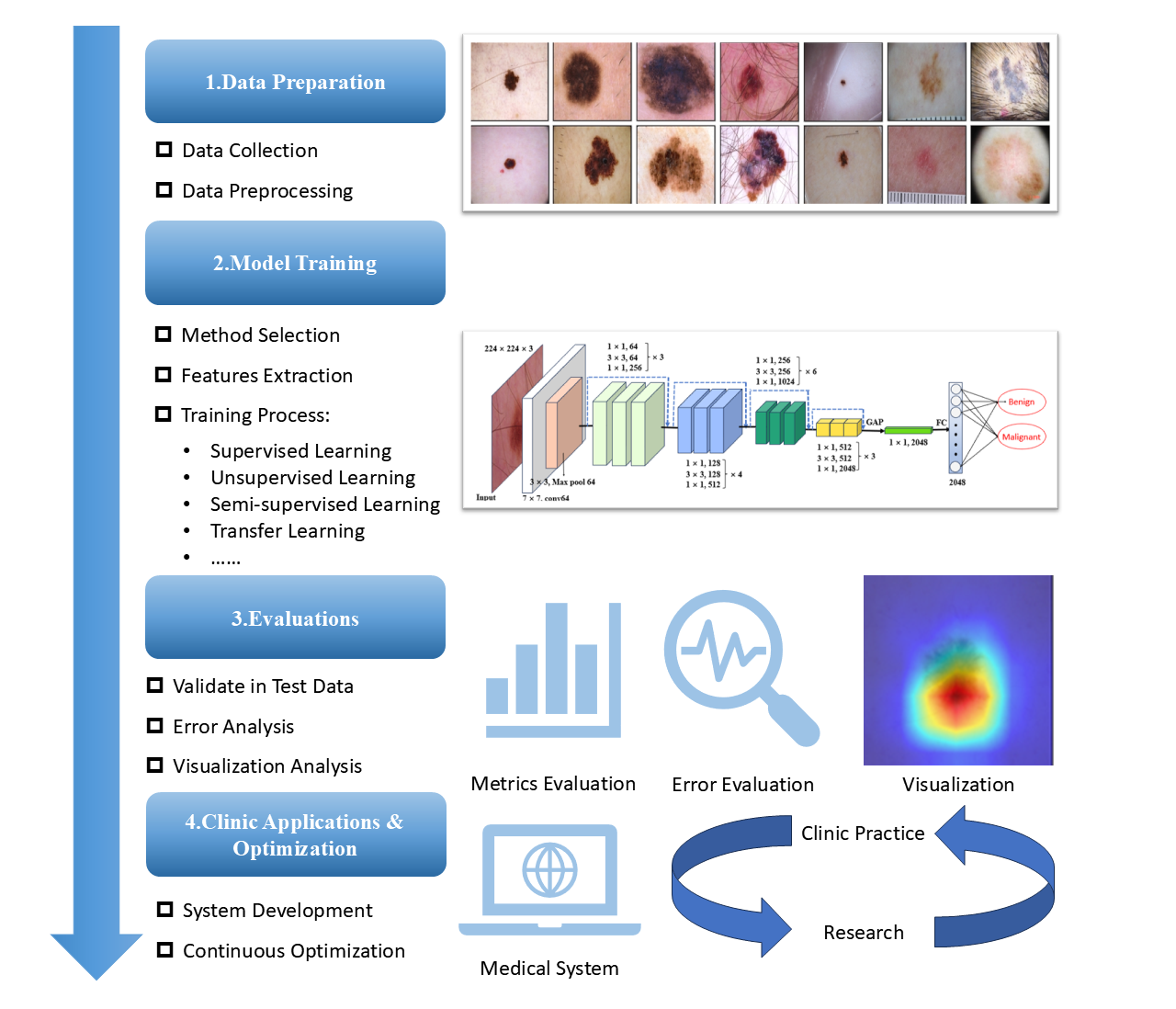}
\caption{Deep Learning Process Flow for Skin Cancer diagnosis.}\label{dl_process}
\end{figure}

To diagnose diseases in the early stage, techniques to analyze medical data, either clinical records or medical images, are fundamentally required \cite{H. K. Bhuyan}. Compared to traditional methods, deep learning-based diagnostic tools offer automated feature extraction, improved classification accuracy, and enhanced efficiency, making them a promising alternative for modern dermatological assessment. The application of deep learning in dermatological diagnosis has significantly advanced the field of automated skin disease detection, offering high accuracy and efficiency in medical image analysis \cite{D. Painuli}. The diagnostic pipeline typically consists of four primary stages as demonstrated in Fig.\ref{dl_process}.

High-quality data is fundamental to deep learning, particularly in medical diagnostic applications \cite{R. Guetari}. High-resolution dermoscopic images for study could be collected from widely recognized benchmark datasets such as ISIC, HAM10000, and Derm7pt, etc. Given the challenges posed by real-world clinical settings, various preprocessing techniques are employed to enhance data quality, including image normalization, hair artifact removal, color correction, contrast enhancement, and data augmentation. These steps mitigate the impact of image noise, intra-class variability, and inter-class similarity, thereby improving model generalization.

Following data preprocessing, the next phase involves training a deep learning model to classify skin diseases accurately. For this step, a correct selection of a deep learning method is needed. Architectures such as Convolutional Neural Networks (CNNs), Vision Transformers (ViTs), and hybrid models could be chosen based on their characteristics in medical imaging tasks. Then, the network is to capture hierarchical representations of skin lesions. For the training process, supervised learning, where models are trained on labeled datasets to perform classification tasks, is a common option \cite{S. Uddin}. Meanwhile, there are also additional choices. Unsupervised learning facilitates pattern recognition without predefined labels, which could also be efficient in diagnosing other diseases \cite{A. A. Patel}. Semi-supervised learning leverages both labeled and unlabeled data to enhance model robustness, which improves model accuracy for maintaining the topological structures to solve the multiple frameworks \cite{R. Sheikhpour}. In transfer learning, pre-trained networks are fine-tuned for skin disease diagnosis, reducing computational cost and improving convergence speed \cite{A. T. Ibrahim}.

The trained model undergoes rigorous evaluation to assess its diagnostic accuracy, sensitivity, specificity, and F1-score using independent test sets. The metrics evaluation ensures the model's ability to generalize beyond the training dataset \cite{I. Ahmad}. To identify misclassification patterns and potential biases in the dataset, particularly in cases of rare or underrepresented skin conditions, error analysis is commonly conducted \cite{T.-W. Wang}. To improve interpretability, visualization techniques are utilized for further analysis. For instance, image regions that influence model predictions can be highlighted via Gradient-weighted Class Activation Mapping (Grad-CAM) so that the difficulty of “black-box” can be overcome. Once validated, deep learning models are integrated into Clinical Decision Support Systems (CDSS) to assist dermatologists in real-time diagnosis, risk stratification, and treatment planning. These systems undergo continuous refinement through real-world clinical feedback loops, ensuring adaptability to new cases, emerging disease patterns, and evolving imaging standards. Furthermore, ongoing model optimization involves incremental learning, dataset expansion, and explainability improvements, enhancing clinical trust and usability \cite{Q. Huang}.

The integration of deep learning into dermatological diagnostics holds transformative potential in improving diagnostic accuracy, reducing interobserver variability, and facilitating early detection of malignant skin lesions. However, deep learning methods still encounter significant obstacles inherited from traditional CAD, including complex features, intra-class variation, and inter-class similarity. Meanwhile, as described above, big data is critical for deep learning methods. In this situation, imbalanced classes of a dataset are a new problem to be considered.

\subsection{Training hyperparameters}

\begin{table}[h]
\centering
\captionsetup{justification=centering,singlelinecheck=false} 
\caption{Summary of Training Hyperparameters.}
\label{Hyperparameters_table}

\begin{tabular}{
    >{\centering\arraybackslash}m{3cm}   
    >{\centering\arraybackslash}m{3cm}   
    >{\centering\arraybackslash}m{8cm}   
}
\toprule
\textbf{Hyperparameter} & \textbf{Example of Value} & \textbf{Purpose} \\ 
\midrule
Optimizer      & Adam   & Stable and adaptive weight updates \\ 
Batch Size     & 32     & Balance between memory usage and gradient stability \\ 
Learning Rate  & 0.001  & Controls update magnitude \\ 
Scheduler      & StepLR & Gradual LR decay for stable late-stage training \\ 
Epochs         & 50     & Sufficient model convergence \\ 
Weight Decay   & 1e-5   & Prevents overfitting through regularization \\ 
\bottomrule
\end{tabular}

\end{table}
Training hyperparameters play a crucial role in shaping model convergence, stability, and generalization. Commonly used hyperparameters are listed in Table \ref{Hyperparameters_table}, which includes the optimizer, such as Adam and SGD, which determines how model weights are updated, and the learning rate and its scheduler, which control the step size and its decay during training. Batch size and number of epochs define how data are processed and how long the model is trained, while regularization strategies such as weight decay, dropout rates, or momentum terms help prevent overfitting and improve robustness. Additional settings, including initialization methods, gradient clipping thresholds, and early-stopping criteria, may further influence optimization behavior. Together, these hyperparameters govern the overall training dynamics and are essential for achieving stable and high-performing deep learning models.

\subsection{Methods to solve complex features}
\label{methods_complex_features}

The complexity of dermatological image features arises from blurred lesion boundaries, irregular shapes, and diverse color variations, etc. Lesion appearances vary significantly across different patients, making it challenging to establish a standardized pattern for classification. Many lesions exhibit unclear demarcation from healthy skin, atypical morphological structures, and heterogeneous pigmentation, further complicating feature extraction. These factors increase the difficulty for deep learning models to accurately capture and differentiate key characteristics, ultimately affecting the reliability of automated skin disease diagnosis. Recently, several studies try to solve the problem, mainly as shown in Table ~\ref{tab:complex_features}.

Several studies focused on identifying the edges of a lesion by analyzing contrasts, gradients, and texture differences in the image. For this, some segmentation techniques are involved. Adegun et al. propose a fully convolutional network (FCN)-based deep learning system enhanced with a probabilistic model. This model refines lesion borders by employing a mean-field approximate probabilistic inference approach combined with a fully connected conditional random field (CRF) using a Gaussian kernel. The framework improves segmentation accuracy by reducing boundary ambiguity and enhancing feature extraction. The proposed method is evaluated on publicly available ISBI 2017 and PH2 datasets, achieving an impressive 98\% accuracy \cite{Adegun2021}. Deepa et al. propose a novel adaptive-based segmentation technique that enhances image quality and improves lesion boundary detection. Their approach involves pre-processing skin images using filtering methods and Adaptive Contrast Limited Adaptive Histogram Equalization (CLAHE), with optimized threshold values determined by the Hybrid Rat Electric Fish Swarm Optimization (HREFSO) algorithm. For segmentation, they introduce the Adaptive Boundary-aware Transformer with Gated Attention Mechanism (ABT-GAM), which enhances boundary delineation and segmentation precision. By optimally tuning segmentation parameters using HREFSO, their method achieves higher segmentation accuracy, demonstrating superior performance in diagnosing skin cancer compared to existing techniques \cite{Deepa2023}. Ahmad et al. propose a dual-phase deep learning pipeline ensembled with Deeplabv3+ for lesion segmentation and a ViT for classification. Meanwhile, an encoder-decoder CNN model is also included in the model. By this method, global relationships across patches are captured by ViT, and lesions, despite noisy and artifact-heavy images, are accurately segmented by Deeplabv3+ \cite{Ahmad2024}. An optimization-driven model incorporating deep learning techniques was proposed by Kumar et al. to overcome the segmentation errors caused by irregular lesion boundaries. Their approach includes Gaussian filtering and Region of Interest (ROI) extraction for noise removal, followed by U-RP-Net, a hybrid segmentation model combining U-Net and RP-Net. The outputs from both networks are fused using a Jaccard similarity-based fusion model for improved segmentation precision. After segmentation, data augmentation is applied to enhance generalization, and skin cancer detection is performed using SqueezeNet, which is optimized with the newly proposed Aquila Whale Optimization (AWO) algorithm, combining Aquila Optimizer (AO) and Whale Optimization Algorithm (WOA) \cite{Kumar2022}.

\begin{longtable}{%
>{\centering\arraybackslash}m{1.5cm}  
>{\centering\arraybackslash}m{1.3cm}    
>{\centering\arraybackslash}m{7cm}    
>{\centering\arraybackslash}m{3cm}    
>{\centering\arraybackslash}m{2.5cm}} 
\caption{Included studies mainly to overcome complex features.} \label{tab:complex_features} \\
\toprule
\textbf{Reference} & \textbf{Year} & \textbf{Method} & \textbf{Datasets} & \textbf{Results} \\
\midrule
\endfirsthead

\multicolumn{5}{c}{{\bfseries Table \thetable\ (continued)}} \\
\toprule
\textbf{Reference} & \textbf{Year} & \textbf{Method} & \textbf{Datasets} & \textbf{Results} \\
\midrule
\endhead

\bottomrule
\endfoot

\cite{Adegun2021} & 2021 & FCN-based deep learning system enhanced with a probabilistic model & ISBI 2017, PH2 & 98\% accuracy \\
\cite{Albraikan2022} & 2022 & ADL-MDC model incorporating CapsNet, Adagrad optimization, k-means clustering, CSO, and SAE & ISIC dataset & 98.27\% accuracy \\
\cite{Kumar2022} & 2022 & Hybrid U-RP-Net segmentation, SqueezeNet with Aquila Whale Optimization (AWO) & ISIC dataset & 92.5\% accuracy \\
\cite{Rasheed2022} & 2022 & Hybrid deep model (handcrafted + CNN features + ReliefF + SVM) & EIR & 88.29\% accuracy \\
\cite{Deepa2023} & 2023 & Filtering methods and Adaptive CLAHE with HREFSO, ABT-GAM & PH2, ISIC 2016, ISIC 2018 & 96.28\% accuracy \\
\cite{Huo2023} & 2023 & HiFuse Network (Global-Local Fusion) with W-MSA + HFF Block + IRMLP & ISIC-2018 & 85.85\% accuracy \\
\cite{Jasil2023} & 2023 & Densenet + ResNet hybrid CNN with multi-task learning & HAM10000 & 95\% accuracy \\
\cite{Sorour2023} & 2023 & Cascaded YOLO-V5 with CGAN data augmentation & Custom & 97\% accuracy \\
\cite{Karthik2024} & 2024 & Swin Transformer + DenseNet169 + GSDW + ENLA + Focal Loss & HAM10000 & 94.21\% accuracy \\
\cite{Ahmad2024} & 2024 & Deeplabv3+ and Vision Transformer (ViT) pipeline & ISIC-2016, ISIC-2017, ISIC-2018, PH2 & 97.73\% accuracy \\
\cite{Prasad2024} & 2024 & ADTBO-SqueezeNet with dual segmentation, feature fusion, and optimization & ISIC dataset & 97.6\% accuracy \\
\cite{Chyad2025} & 2025 & Adversarial machine learning (AML), GANs, CNN-RNN hybrid, multi-color space feature fusion & Custom & 99\% accuracy \\
\cite{Ozdemir2025} & 2025 & ConvNeXtV2 + separable self-attention hybrid deep learning & ISIC-2019 & 93.48\% accuracy \\
\cite{Mehidi2025} & 2025 & Combined Xception and Inception modules & ArsenicSkinImageBD & 97.69\% accuracy \\
\cite{Dillshad2025} & 2025 & Residual blocks optimization with entropy-controlled Euclidean Distance & ISIC-2019 & 99.3\% accuracy \\
\cite{Rashid2025} & 2025 & 3D ResNet50 with transfer learning & ISIC-2024 & 93.88\% accuracy \\
\cite{Ozdemir2025b} & 2025 & ConvNeXtV2 blocks and focal self-attention & ISIC-2019 & 93.60\% accuracy \\
\cite{Sargunan2025} & 2025 & DLNDP feature extraction and R2U-Net segmentation & Custom & 90.42\% accuracy \\
\end{longtable}

Combined with multiple architectures, hybrid models could be created to capture local and global features, so that complex features can be solved. Jasil et al. integrated a DenseNet and a ResNet into a hybrid CNN model. Herein, feature representations by establishing dense connectivity between layers are abstracted by DenseNet, while the deep feature extraction through skip connections is enhanced by ResNet. Multi-scale feature aggregation is applied to ensure the model captures both local and global lesion features. After that, multi-task learning is performed. The segmentation task isolates lesion regions, providing precise contour boundaries and the classification task identifies the lesion type. Besides, data augmentation techniques are also involved \cite{Jasil2023}. A dual-track hybrid deep learning system that synergistically combines Swin Transformer and a custom Dense Group Shuffle Non-Local Attention (DGSNLA) Network is designed by Karthik et al. to address the limitations of conventional CNNs that struggle with capturing both local fine-grained details and global contextual features. In this study, Swin Transformer is used to extract global dependencies through window-based multi-head self-attention, enabling effective modeling of long-range spatial relationships in dermoscopic images. DGSNLA Network is used for efficient local feature extraction and cross-channel information exchange. To improve contextual awareness, Enhanced Non-Local Attention is integrated to model non-local correlations. Moreover, focal loss is also applied to deal with imbalanced data \cite{Karthik2024}. To overcome the issues of low contrast, image noise, irregular lesion boundaries, and feature extraction inefficiencies, Albraikan et al. developed the Automated Deep Learning-Based Melanoma Detection and Classification (ADL-MDC) model, which integrates multiple computational techniques for enhanced diagnostic accuracy. The ADL-MDC framework includes contrast enhancement and data augmentation for pre-processing, k-means clustering for segmentation, Adagrad-optimized Capsule Network (CapsNet) for feature extraction, and Crow Search Optimization (CSO) with Sparse Autoencoder (SAE) for classification. By leveraging these advanced methods, the model improves robustness in handling feature complexity, enhances segmentation precision, and optimizes classification performance. Experimental validation on benchmark datasets (ISIC dataset) demonstrates the superior accuracy and reliability of the proposed method, achieving high classification metrics \cite{Albraikan2022}. Chyad et al. established a deep learning framework integrating adversarial machine learning (AML) and generative adversarial networks (GANs), utilizing a dataset with 17 million patches across multiple color spaces (RGB, YCbCr, HSV, YUV). Their method includes feature fusion, adversarial training, and the application of hybrid CNN-RNN architectures to improve skin lesion classification accuracy \cite{Chyad2025}. Mehidi et al. introduces ArsenicNet, a deep learning model combining Xception and Inception modules to detect arsenic-induced skin disease from images \cite{Mehidi2025}. Sargunan et al. established ensembled with image preprocessing with a Gaussian filter, feature extraction using a novel Discrete Local Neighborhood Difference Pattern (DLNDP) and shape features, followed by segmentation via R2U-Net \cite{Sargunan2025}.

For the same aim of creating hybrid models, multi-scale feature extraction and fusion is applied. Rasheed et al. proposed a hybrid deep neural network for automatic classification of eczema. The method integrates handcrafted features and deep activated features optimized by ReliefF and PSO algorithms, and classified based on SVM. Data augmentation and GANs are also used to cope with class imbalance \cite{Rasheed2022}. Hierarchical Multi-Scale Feature Fusion Network (HiFuse) is put forward by Huo et al., which utilized a Windows Multi-head Self-Attention (W-MSA) mechanism, inspired by Swin Transformer, to capture long-range dependencies. For preserving fine-grained spatial details, a depthwise convolution is utilized. Then, a Hierarchical Feature Fusion (HFF) Block merges global and local representations for prediction. The HFF consists of spatial attention, channel attention, and an Inverted Residual Multi-Layer Perceptron (IRMLP), which ensures a smooth transition between different hierarchical levels, enhancing multi-scale feature integration \cite{Huo2023}. Sorour et al. utilized the YOLO-V5 object detection algorithm to detect and localize melanoma and vitiligo. The classification and localization are based on cascaded YOLO-V5 models, where the output weights from one model are used to initialize the next. In this study, complex features were tackled by YOLO-V5’s multi-scale detection and CSPDarknet backbone, where the denoising layers and CGAN-generated realistic images for augmentation were also involved in this article \cite{Sorour2023}. Prasad et al. put forward a skin cancer detection framework that integrates Aquila Driving Training-Based Optimization (ADTBO) with SqueezeNet. The method utilizes a Minkowski-based dual network, including PsiNet and DeepJoint segmentation for precise lesion segmentation, followed by data augmentation and comprehensive feature extraction. The SqueezeNet model’s weights are optimized by ADTBO, a hybrid of Aquila Optimizer and Driving Training-Based Optimization, to enhance learning performance \cite{Prasad2024}. Ozdemir et al. designed an innovative hybrid deep learning model that combines ConvNeXtV2 blocks with separable self-attention mechanisms for multiclass skin cancer classification. The ConvNeXtV2 blocks are used in the early stages to capture fine-grained local features. Meanwhile, the separable self-attention layers are utilized in the later stages, efficiently modeling global context with reduced computational complexity. Then, to handle the class imbalance, data augmentation, transfer learning, and a weighted loss function are adapted \cite{Ozdemir2025}. Dillshad et al. proposed a novel deep learning framework for skin cancer classification that combines two lightweight CNN models with residual blocks optimized via Bayesian tuning. This method also fused deep features based on entropy-controlled Euclidean Distance method \cite{Dillshad2025}. Rashid et al. enhanced the 3D ResNet50 model using transfer learning for accurate skin cancer classification \cite{Rashid2025}. Ozdemir et al. integrated ConvNeXtV2 blocks and focal self-attention to effectively extracts local features and highlights critical lesion regions \cite{Ozdemir2025b}.

\subsection{Methods to solve image noise}
\label{methods_image_noise}

The presence of significant image noise poses challenges in extracting key diagnostic features from dermoscopic images. In skin nevus datasets, critical lesion information constitutes only a small proportion of the overall image data, making it difficult to isolate and analyze. For instance, studies have proven that an artifact scale bar will distort CNN outputs by diverting attention from lesions to irrelevant image regions~\cite{Winkler2021}. Noise could also occur due to multiple unpredictable factors, such as the age of patients, slide origin, and even the scanner type~\cite{Schmitt2021}. Additionally, the visibility of these key features is often compromised due to visual artifacts, including other skin spots, hair occlusions, and variations in lighting conditions, further complicating accurate feature extraction and classification. To address the image noise, studies have been conducted as shown in Table~\ref{tab:image_noise}.

To handle image noise, noise removal and several preprocessing technologies are highlighted in recent studies. To classify skin cancer lesions into benign and malignant classes while minimizing training time and space complexity, Montaha et al. developed a lightweight convolutional neural network named SCNN\_12 and applied comprehensive image preprocessing, segmentation, and a down-scaling method using box blur and bilateral filtering~\cite{Montaha2022}. Reshma et al. developed an Intelligent Multilevel Thresholding with Deep Learning (IMLT-DL) model for skin lesion segmentation and classification. The approach begins with Top-hat filtering and inpainting for preprocessing, removing artifacts like hair and enhancing image quality. Next, Mayfly Optimization (MFO) combined with Kapur’s multilevel thresholding segments the lesion areas. Features are extracted using InceptionV3, and classification is performed with a Gradient Boosted Tree (GBT) model~\cite{Reshma2022}. Salma et al. utilized morphological filtering for hair and artifact removal. Then, they applied a Grab-cut segmentation in the HSV color space. After that, the ABCD feature extraction is used to simplify complex lesion characteristics. Data augmentation is also used to balance the dataset~\cite{Salma2022}. Yousefi et al. propose an Automated Melanoma Detection (AMD) mechanism, which integrates Discrete Cosine Transform (DCT) features and metadata to enhance accuracy and sensitivity. A removal of hair artifacts and patient metadata, followed by image standardization, is performed in this study to ensure high-resolution analysis. In feature extraction, the Radon transform is applied to eliminate white noise, followed by Top-hat filtering to remove irrelevant background data. Discrete Wavelet Transform (DWT) and Discrete Cosine Transform (DCT) are then used to extract robust frequency domain features. Then, Locality Sensitive Discriminant Analysis (LSDA) is employed to reduce the number of extracted features while preserving discriminative information. For classification, processed images are divided into training and test datasets. The best-performing model is selected using Decision Trees (DT), K-Nearest Neighbors (K-NN), and Support Vector Machines (SVM), with DT emerging as the best classifier. Additionally, patient metadata, including sex, rash site, and age, is incorporated to enhance classification performance~\cite{Yousefi2024}.

Technologies such as data augmentation and contrast enhancement could also work against image noise by preprocessing noise and increasing the diversity of training data. Based on the investigation of how background color information in clinical dermatological images affects the performance of deep learning models in skin disease classification, Zhou et al. proposed a schema involving background masking, which involves white, black, blue, and green colors. Data augmentation is also used in this work. It was found that models trained on images with a green background consistently achieved the highest performance~\cite{Zhou2022}. Under the challenge of effectively handling high-dimensional and redundant features in skin lesion classification, BF2SkNet offers a robust solution by leveraging deep feature fusion and optimized feature selection. This framework combines Inception-ResNetV2 and NasNet-Mobile to extract complementary feature representations, which are then refined using a fuzzy entropy-based Slime Mould Algorithm (Entropy-SMA) to retain only the most discriminative elements. To further enhance the model’s performance, selected features are fused via a Serial-Threshold strategy and passed to conventional machine learning classifiers for final prediction. Moreover, extensive data augmentation is employed to address class imbalance. Besides, contrast-based data augmentation is used for solving data imbalance~\cite{BF2SkNet2023}. Integrated with deep learning and an optimized feature selection method, Dillshad et al. present D2LFS2Net, an end-to-end framework for multi-class skin lesion classification. The system employs a hybrid contrast enhancement for better lesion visibility and uses MobileNetV2 and NASNet-Mobile for feature extraction via deep transfer learning. Extracted features are fused using a dual-threshold serial approach, followed by variance-controlled Marine Predator Optimization (VcMPA) to select the most discriminative features. The final classification is performed using machine learning classifiers, with Cubic SVM achieving the best results~\cite{Dillshad2023}. Adepu et al. proposed a Knowledge Distillation (KD)-based lightweight Deep CNN framework, coupled with Cost-Sensitive Learning (CSL) using Focal Loss (FL) to improve classification sensitivity. Their approach includes artifact removal using an inpainting algorithm, as well as data augmentation techniques such as Sprinkled and Microscopic CutOut, which help regularize the model and mitigate overfitting. The teacher-student model setup uses EfficientNet-B5 as the teacher and EfficientNet-B2 as the student, where the teacher model’s soft targets are used to guide the student model, improving feature discrimination in melanoma classification~\cite{Adepu2023}. Rokade et al. conducted a Blockchain-based Deep Learning System (TCIO-LeNet) that integrates secure data management with optimized deep learning for robust skin disease detection. In this research, skin images and patient data are stored in a blockchain ledger, ensuring tamper-proof, decentralized, and secure data sharing in medical applications. In pre-processing, Adaptive Bilateral Filtering (ABF) is applied to reduce noise and improve image quality by removing artifacts, hair, and uneven lighting conditions. Then, a modified DeepJoint segmentation technique is introduced, incorporating Kumar-Hassebrooks similarity measures to enhance lesion extraction accuracy. The model extracts texture, statistical, and local features and applies various augmentation techniques to improve model generalization. Finally, the Transit Circle Inspired Optimization (TCIO) algorithm is integrated with LeNet, modifying its weights for improved feature learning and classification accuracy~\cite{Rokade2024}. AbuAlkebash established a robust and explainable skin cancer detection framework that integrates YOLOv8 for lesion localization and ViT for classification, which also enhanced model interpretability using Grad-CAM and SHAP~\cite{AbuAlkebash2025}.

Not only are hybrid models and feature fusion efficient in dealing with complex features, but they also have a critical impact on coping with image noise. To realize this, technologies including optimized loss function, attention mechanisms, and transfer learning, etc. are introduced. For the segmentation of telangiectasia, which is characterized by serpiginous and arborizing vessels and a key visual feature for BCC diagnosis, Maurya adopts a U-Net architecture optimized with hybrid loss functions and trains it on BCC dermoscopic images to automatically detect and segment blood vessels~\cite{Maurya2022}. To guide the network’s focus towards the most relevant image features, Pacheco and Krohling introduce MetaBlock, an attention-based mechanism based on metadata. The MetaBlock allows dynamic adjustment of feature importance, improving both performance and interpretability. Weighted loss functions and data augmentation are also adapted to deal with imbalanced classes~\cite{Pacheco2021}. A novel framework for multiclass skin lesion recognition using deep learning and explainable AI is put forward in Ahmad et al.’s research, which includes two optimized Xception and ShuffleNet models via transfer learning. Deep features are extracted from both models and fused to enhance classification performance. To address computational complexity, an improved Butterfly Optimization Algorithm (BOA) is introduced for optimal feature selection, followed by classification using various machine learning classifiers. Besides, data augmentation is also conducted~\cite{Ahmad2023}. The effect of deep transfer learning for the binary classification of uveal melanoma (UM) from fundus images, focusing on differentiating choroidal nevus (CN) from non-lesion cases, is explored by Shakeri et al. Herein, 4 pre-trained CNN models are trained on a manually curated dataset of 854 fundus images. To address the black-box nature of deep learning, Moreover, Shapley Additive exPlanations (SHAP) analysis was used to interpret model predictions by highlighting image regions contributing most to classification decisions. Class weighting also helped manage class imbalance in this work~\cite{Shakeri2023}. Saeed et al. proposed a hybrid framework with VGG19 and an SVM model. In this model, GAN and ESRGAN are integrated, where the ESRGAN is used to upscale and enhance image resolution, and GAN helps improve image sharpness, respectively~\cite{Saeed2023}. Verma et al. proposed a hybrid deep learning and machine learning ensemble approach for improving classification accuracy in imbalanced skin cancer datasets. Their method integrates feature extraction from pre-trained deep learning models (DenseNet201, Xception, MobileNet), feature concatenation, and ensemble machine learning classifiers to enhance robustness against noisy and distorted images. To further reduce noise, the study employs image pre-processing techniques, including RGB to grayscale conversion, black-hat transformation, binary Otsu thresholding, and inpainting to remove extraneous noise and enhance lesion visibility. Additionally, data augmentation techniques are also used~\cite{Verma2024}. To enhance feature extraction, Ntampakis et al. developed a hybrid deep learning framework that integrates self-attention fusion and Vision Transformers (ViTs) for skin lesion classification. The study employs adaptive contrast enhancement, noise filtering, and artifact removal techniques to improve the visibility of lesion boundaries. A novel self-attention-based feature fusion is introduced to integrate multi-scale features and enhance lesion representation, making the model more robust to noisy and occluded regions. The processed images are fed into a ViT-based classifier, which leverages global attention mechanisms to improve the recognition of complex lesion patterns and mitigate the impact of noise-related distortions~\cite{Ntampakis2024}. Zareen et al. combined Convolutional Neural Networks (CNN) and Recurrent Neural Networks (RNN) with a ResNet-50 backbone and Long Short-Term Memory (LSTM) units. This model effectively integrates spatial feature extraction and temporal sequence learning to enhance classification accuracy for various skin cancer types. Where, ResNet-50 is employed to extract spatial features, capturing complex patterns in lesion images. LSTM layers are incorporated to analyze sequential dependencies and lesion progression, improving robustness against variations in lesion appearance. The method researched 99.06\% accuracy in a dataset constructed based on ISIC~\cite{Zareen2024}. Midasala et al. introduced a multilevel feature extraction (MFE) approach integrated with an unsupervised learning (USL) segmentation method to improve lesion detection and classification accuracy. The MFEUsLNet, an unsupervised learning-based deep learning model is designed in this study. In pre-processing, a bilateral filter is used to remove noise artifacts and enhance lesion visibility. Then, The USL-KMC (unsupervised K-means clustering) algorithm is applied for precise lesion segmentation. After that, a combination of Gray Level Co-occurrence Matrix (GLCM) for texture extraction and Redundant Discrete Wavelet Transform (RDWT) for color and spatial feature extraction is used. Finally, a RNN is trained on these multilevel extracted features to classify seven types of skin cancer. The experiment is conducted on ISIC-2020, metrics such as accuracy, precision, and specificity are all higher than 99\%~\cite{Midasala2024}. Girmaw et al. established a deep learning-based classification system using transfer learning with three pre-trained models: EfficientNetB7, MobileNetV2, and DenseNet201. The system is designed to enhance feature extraction, improve classification accuracy, and provide real-time detection capabilities. Besides common data augmentation such as rotation, flipping, and zooming, this study utilizes EfficientNetB7, MobileNetV2, and DenseNet201 for automated feature extraction, leveraging their strengths in handling complex textures and patterns. A modified SoftMax classifier is employed to categorize skin diseases into four classes. Performing animal skin disease as the study did, there are still some reference values for human skin disease research, become the characteristic of image noise is the same~\cite{Girmaw2025}.

Furthermore, several literature aforementioned paid attention to the image noise synchronously~\cite{Deepa2023,Kumar2022}. The obstacles of complex features and image noise might occur simultaneously. The optimization of hyperparameters is critical for the improvement of the models’ capacity to deal with complex features integrated with image noise. However, traditional CNN-based models often select hyperparameters arbitrarily, leading to suboptimal performance. Thus, Shaik et al. aimed to systematically fine-tune hyperparameters. Based on the constructed DCNN, iterative tuning is performed, and Adam optimizer with a learning rate of 0.1 and dropout rate of 0.2 is found to yield the best results. In this study, bilateral filtering is utilized to remove image noise while preserving edges, and data augmentation is also adapted~\cite{Shaik2024}.

\begin{longtable}{%
>{\centering\arraybackslash}m{1.5cm}   
>{\centering\arraybackslash}m{1.3cm} 
>{\centering\arraybackslash}m{7cm}   
>{\centering\arraybackslash}m{3cm}   
>{\centering\arraybackslash}m{2.5cm}}  
\caption{Studies primarily to overcome image noise.}
\label{tab:image_noise} \\
\toprule
\textbf{Reference} & \textbf{Year} & \textbf{Method} & \textbf{Datasets} & \textbf{Results} \\
\midrule
\endfirsthead

\multicolumn{5}{c}{{\bfseries Table \thetable\ (continued)}} \\
\toprule
\textbf{Reference} & \textbf{Year} & \textbf{Method} & \textbf{Datasets} & \textbf{Results} \\
\midrule
\endhead

\bottomrule
\endfoot

\cite{Pacheco2021} & 2021 & MetaBlock (attention-based image and metadata fusion) & ISIC-2019, PAD-UFES-20 & 77\% BACC \\
\cite{Salma2022} & 2022 & Grab-cut, ABCD features simplification, data augmentation & ISIC-2017, HAM10000 & 99.87\% accuracy \\
\cite{Maurya2022} & 2022 & U-Net-based vessel segmentation with hybrid loss & HAM10000 and NIH & 37.8\% Jaccard Index \\
\cite{Montaha2022} & 2022 & Shallow CNN (SCNN\_12) with image preprocessing and down-scaling & ISIC dataset & 98.87\% accuracy \\
\cite{Zhou2022} & 2022 & Background masking (color-based) + ResNet50 + data augmentation & Custom & Up to 5.47\% F1-score improvement with green background \\
\cite{BF2SkNet2023} & 2022 & BF2SkNet: Deep feature fusion (InceptionResNetV2 + NasNet) + Entropy-SMA selection & HAM10000, ISIC-2018 & 97.1\% accuracy \\
\cite{Reshma2022} & 2022 & IMLT-DL: MFO + Kapur’s segmentation + InceptionV3 + GBT & ISIC dataset & 99.2\% accuracy \\
\cite{Adepu2023} & 2023 & Knowledge Distillation (KD), Cost-Sensitive Learning (CSL) with Focal Loss (FL), Inpainting-based artifact removal, EfficientNet-B5 teacher, EfficientNet-B2 student & ISIC-2020 & 0.9295 AUC \\
\cite{Ahmad2023} & 2023 & Deep learning with Xception and ShuffleNet + feature fusion + BOA feature selection & HAM10000 \& ISIC-2018 & 99.3\% accuracy \\
\cite{Dillshad2023} & 2023 & D2LFS2Net: Deep transfer learning + VcMPA optimization + feature fusion & HAM10000 & 94.4\% accuracy \\
\cite{Shakeri2023} & 2023 & DenseNet169 + SHAP analysis + transfer learning & Custom & 89\% accuracy \\
\cite{Saeed2023} & 2023 & VGG19 + SVM + GAN + ESRGAN & ISIC-2019, ISIC-2020 & 93\% accuracy \\
\cite{Verma2024} & 2024 & Hybrid deep learning and machine learning ensemble (DenseNet201, Xception, MobileNet) with pre-processing techniques (grayscale conversion, black-hat transform, Otsu thresholding, inpainting) & HAM10000 \& ISIC datasets & 98.7\% accuracy \\
\cite{Ntampakis2024} & 2024 & Self-Attention Fusion + ViT for Skin Lesion Classification & HAM10000 \& ISIC datasets & 97.8\% accuracy \\
\cite{Rokade2024} & 2024 & Blockchain-based deep learning system (TCIO-LeNet) with modified DeepJoint segmentation and TCIO optimization & HAM10000 and ISIC-2019 & 92.5\% accuracy \\
\cite{Zareen2024} & 2024 & Hybrid CNN-RNN model with ResNet-50 and LSTM for skin cancer classification & ISIC datasets & 99.06\% accuracy \\
\cite{Midasala2024} & 2024 & MFEUsLNet (Multilevel Feature Extraction + Unsupervised Learning) with USL-KMC segmentation, GLCM \& RDWT feature extraction, and RNN classifier & ISIC-2020 & 99.179\% accuracy \& 99.824\% Recall \\
\cite{Yousefi2024} & 2024 & Automated Melanoma Detection (AMD) with DCT features, Radon transform, DWT, LSDA, and metadata integration & ISIC-2020 & 70.0\% accuracy \& 68.0\% F1-score \& 71.0\% precision \\
\cite{Shaik2024} & 2024 & DCNN with iterative optimized hyperparameters & HAM10000 & 96\% accuracy \\
\cite{Girmaw2025} & 2025 & Deep learning-based classification using transfer learning & Custom & 99.01\% accuracy \\
\cite{AbuAlkebash2025} & 2025 & Yolo, Grad-CAM \& SHAP & Custom & 93\% F1-score \\
\end{longtable}

Recent studies also show that noise handling should be accompanied by explicit modeling of prediction reliability. Deep models often give confident scores even when the input image contains artefacts or distribution shifts. This reduces clinical safety. Uncertainty estimation provides a way to identify low-confidence results. Common methods include Monte Carlo dropout, deep ensembles and evidential learning. These methods help separate model uncertainty and data uncertainty. Calibration is also important because raw softmax outputs usually do not reflect true confidence. Several technologies adapted this technology, such as Swin-T, Meta-UNet, Quad-Step Approach and multi-stage multi-modal learning. In the calibration model, temperature scaling and other simple post-training techniques can improve probability calibration. Reliability diagrams and Brier scores are often used to assess calibration quality. Integrating uncertainty estimation with noise reduction can improve the trustworthiness of skin lesion analysis and support safer clinical use.

\subsection{Methods to solve intra-class variation and inter-class similarity}
\label{Methods to solve intra-class variation and inter-class similarity}

Intra-class variation and inter-class similarity could also be significant obstacles to automatic skin disease diagnosis, which causes fuzzy category distinction~\cite{Mirikharaji2023}. Intra-class variation refers to the significant differences among images within the same category, meaning that skin lesions of the same type can appear highly diverse in terms of color, texture, size, and location. This makes it challenging for models to generalize and correctly classify all variations of a given skin disease. Inter-class similarity, on the other hand, describes the high degree of visual resemblance between different categories, such as benign and malignant skin lesions, which often share similar color distributions, textures, and structural patterns. This overlap in appearance reduces the distinctiveness between classes, making accurate differentiation more difficult~\cite{Ren2020}. This problem put forward a tough challenge and attracted various researchers to address. Meanwhile, this type of obstacle is also discussed with complex features. The included studies, primarily to cope with intra-class variation and inter-class similarity, are listed in Table~\ref{tab:intra_inter_variation}.

Hybrid and ensemble learning approaches are equally accessible to deal with the challenge of intra-class variation and inter-class similarity. Ahmed et al. built an ensemble learning framework combined with Xception, Inception-ResNet-V2, and NasNetLarge. These models are fused using a LightGBM classifier, an efficient gradient-boosting decision tree framework that helps in combining the features from the CNNs~\cite{Ahmed2020}. Song presents an end-to-end multi-task deep learning framework for automatic skin lesion analysis, capable of performing classification, detection, and segmentation simultaneously on dermoscopy images. The architecture incorporates a Feature Pyramid Network (FPN), Region Proposal Network (RPN), and three convolutional subnets. A combined loss function based on focal loss and Jaccard distance is introduced to mitigate class imbalance and enhance segmentation accuracy~\cite{Song2020}. Salamaa et al. explored a deep learning-based system based on a hybrid architecture with ResNet-50 and SVM. Wherein, the FC layers are replaced with SVM. Then, a preprocessing pipeline that includes median filtering, contrast enhancement, and edge detection is used to enhance the image quality and make the critical features more obvious. Moreover, data augmentation and transfer learning are applied to counter limited data~\cite{Salamaa2021}. Giovanetti et al. integrated both image and metadata information into an ensemble model to enhance classification accuracy for distinguishing melanoma from non-melanoma lesions. In this study, Image-only CNN model is established for feature extraction from dermoscopic images. Then, they utilized image-and-metadata-based CNN models to incorporate patient-related data~\cite{Giovanetti2022}. Hu et al. propose a hybrid deep residual network and Fisher vector (ResNet-FV) approach to address the challenges of intra-class variation and inter-class similarity in skin lesion classification. In specific, the proposed ResNet-FV framework integrates ResNet-50 for deep feature extraction with Fisher vector encoding to enhance image representation. By leveraging pre-trained ResNet-50 on ImageNet, the model extracts local feature descriptors from dermoscopic images. These extracted features are then aggregated into Fisher vectors, which provide a high-level semantic representation of the lesions, improving classification accuracy. Finally, a Support Vector Machine (SVM) classifier is trained using the Fisher vectors to differentiate between various skin lesion categories~\cite{Hu2022}. Suleiman et al. proposed a two-step hierarchical binary classification model that integrates transfer learning with pre-trained deep learning models such as DenseNet121 and the random forest (RF) algorithm. The two-step classification process first distinguishes between benign lesions and others, followed by a second classification step to differentiate between melanoma and seborrheic keratosis~\cite{Suleiman2024}.

Capturing both local and global features is considerable in dealing with intra-class variation and inter-class similarity, too. The representative technologies including transfer learning, attention mechanisms, and attention-based architectures are widely utilized. To ensure that the network focuses on semantically meaningful regions of the lesion, Qian et al. established a Convolutional Neural Network (CNN) model incorporating Multi-Scale Attention Blocks (GMAB) and Class-Specific Loss Weighting. The GMAB mechanism improves the model’s ability to capture fine-grained details by using multiple scales to extract spatial and channel-wise attention features~\cite{Qian2022}. Hu et al. put forward a multi-scale feature fusion network based on EfficientNetV2, which integrates shallow and deep features to capture both local and global characteristics of skin lesions~\cite{Hu2024}. Suleman et al. improved the MobileNet by leveraging multi-scale feature extraction by employing different kernel sizes and strides across multiple convolutional layers to capture fine-grained details. Thus, the Smart MobiNet is constructed. In this model, attention mechanisms are incorporated into the architecture to enable the model to focus on critical regions of the skin lesion images. Meanwhile, data augmentation is also involved in this study~\cite{Suleman2023}. Tan et al. proposed a dual-branch and global-local attention (GLA) network based on ResNet50, which is called DGLA-ResNet-50. The proposed DGLA-ResNet50 model integrates a Dual-Branch Input Network (DBI) and GLA mechanism to enhance skin lesion recognition by addressing intra-class variation and inter-class similarity. The DBI network extracts image features at different resolutions, expanding the receptive field and improving robustness against scale variations in lesions. Meanwhile, the GLA mechanism consists of Horizontal-Vertical Attention (HVA) for capturing long-range dependencies in both horizontal and vertical dimensions and Local Attention (LA) for extracting fine-grained lesion details, enabling better discrimination of subtle visual patterns. Weighted Random Sampling (WRS) is applied to mitigate class imbalance, ensuring that minority classes receive adequate representation during training~\cite{Tan2024}. Capurro et al. constructed a Transformer-based self-attention mechanism to perform better feature learning across variations and introduced a deep learning-based system for automated direct immunofluorescence (DIF) pattern recognition in autoimmune bullous diseases (AIBDs). Besides, transfer learning is also pre-trained in ImageNet models to mitigate limited data availability and a weighted cross-entropy loss function is applied to address class imbalance. This research involved both the intra-class variation and class imbalance in skin disease classification~\cite{Capurro2024}. Du et al. present a reinforcement learning-based feature selection (RLFS) framework, including an iterative feature selection approach that incorporates deep reinforcement learning (RL) and dynamic data resampling. A deep neural network-based decision network is designed to act as an RL agent and the RL model learns to select the most relevant features iteratively, optimizing for classification accuracy. Before each iteration, a balanced dataset is generated by resampling. Since medical data can vary due to resampling, a GCN-based state encoding method is used to convert it into fixed-length feature representations, making these representations suitable for deep learning models. After training, a weighted model evaluates feature importance across multiple iterations, ensuring that only the most relevant features are retained~\cite{Du2025}. Zhang et al. proposed a DermViT, which simulates the physician's diagnostic process by including multi-scale observation, intelligent focusing and feature screening for skin cancers’ diagnosis~\cite{Zhang2025}. Li et al. proposed a lightweight Local-Global Fusion (LGF) model for skin lesion classification, combining RegNet for local feature extraction and Transformers for global context understanding. Using channel attention and residual connections across four stages, the model effectively integrates pathological information~\cite{Li2025}. Saeed et al. established a dual-path deep learning architecture for accurate skin cancer detection called EG-VAN. This method combined EfficientNetV2S for efficient multi-scale processing with an enhanced ResNet50 using Spatial-Context Group Attention (SCGA) and Non-Local Blocks for fine-grained local and global feature capture~\cite{Saeed2025}.

To improve the performance of machine learning models by eliminating irrelevant or redundant features, feature selection is efficient by selecting a subset of relevant features from a larger set of available features. In the situation of intra-class variation and inter-class similarity, black-box systems, as conventional CNN-based models often function, make it more difficult to interpret the predictions. This makes continuous optimization impossible. Therefore, Nigar et al. integrate Local InterpreTable Model-Agnostic Explanations (LIME) to provide visual justifications for classification decisions, enabling better trust and understanding for dermatologists. The technology is deployed based on a ResNet-18 model. LIME is incorporated to explain individual predictions, which will highlight image regions that contribute most to classification decisions, enabling visual validation of model outputs~\cite{Nigar2022}. Riaz et al. combined deep learning-based high-level feature extraction and handcrafted local texture features to capture global lesion characteristics and preserve fine-grained skin lesion details, respectively. The extracted features from both methods are fused and processed together, creating a robust representation. Then, hair removal and noise reduction techniques are applied to enhance image quality, together with cost-sensitive learning, which is used to balance class distribution, reducing bias toward majority classes~\cite{Riaz2023}. Wang et al. utilized Class Activation Mapping (CAM)-based localization to enhance feature separation across classes to realize a global-lesion-assisted inter-class discrimination learning. For history features enhanced intra-class similarity learning, this work implements a feature updating mechanism that generates a class center based on deep features of all samples in one class and preserves historical feature consistency to improve intra-class feature concentration~\cite{Wang2023}. A deep learning-based system to predict gene expression profiles (GEP) of uveal melanoma directly from digitized cytopathology images is proposed by Liu et al. In their work, automated extraction of over 121,000 ROIs from whole-slide images, which are processed using a ResNet-50 backbone enhanced with dual attention mechanisms, are utilized to capture both spatial and channel-wise feature importance. To transition from ROI-level predictions to slide-level classification, the model employs feature aggregation guided by CAMs to emphasize diagnostically salient regions. After that, data augmentation at the slide level is used to improve generalization. Finally, a two-layer artificial neural network (ANN) classifies the aggregated features into GEP class 1 or class 2, achieving high predictive performance while reducing reliance on costly genetic testing~\cite{Liu2023}. Slian et al. integrated a contextual feature pyramid (CFP) for multi-scale feature extraction to learn different perceptual fields to construct a CFPNet-M architecture. Together with the CFP, the use of attention gates helps the network focus on relevant regions of the image. Meanwhile, the proposed architecture involves a two-step segmentation strategy. The first module pre-segments the entry echo layer, which is the most easily distinguishable in HFUS images, and this information is then used to assist in the segmentation of other layers (SLEB and dermis) in the second module.

\begin{longtable}{%
>{\centering\arraybackslash}m{1.5cm}   
>{\centering\arraybackslash}m{1.3cm} 
>{\centering\arraybackslash}m{7cm}   
>{\centering\arraybackslash}m{3cm}   
>{\centering\arraybackslash}m{2.5cm}}  
\caption{Included studies primarily for intra-class variation and inter-class similarity.}
\label{tab:intra_inter_variation} \\
\toprule
\textbf{Reference} & \textbf{Year} & \textbf{Method} & \textbf{Datasets} & \textbf{Results} \\
\midrule
\endfirsthead

\multicolumn{5}{c}{{\bfseries Table \thetable\ (continued)}} \\
\toprule
\textbf{Reference} & \textbf{Year} & \textbf{Method} & \textbf{Datasets} & \textbf{Results} \\
\midrule
\endhead

\bottomrule
\endfoot

\cite{Ahmed2020} & 2020 & Ensemble Learning & ISIC-2019 & 0.9404 accuracy \\
\cite{Song2020} & 2021 & End-to-end multi-task framework with Focal + Jaccard loss & ISBI-2016, ISIC-2017 & 89.1\% accuracy \\
\cite{Salamaa2021} & 2021 & ResNet50 + preprocessing (median filtering, contrast enhancement, and edge detection) + SVM & ISIC-2017, HAM10000 \& ISBI-2016 & 99.19\% accuracy \\
\cite{Giovanetti2022} & 2022 & Ensemble of Image-Only CNNs and Image+Metadata CNNs & ISIC-2020 & 0.9477 AUC \\
\cite{Hu2022} & 2022 & ResNet-50 + Fisher vector encoding + SVM & ISIC-2018 & 93.0\% accuracy \\
\cite{Nigar2022} & 2022 & ResNet-18-based CNN + LIME & ISIC-2019 & 94.47\% accuracy \\
\cite{Qian2022} & 2022 & CNN-based model with Multi-Scale Attention Blocks (GMAB) and Adaptive Class-Specific Loss Weighting & HAM10000 & 91.6\% accuracy \\
\cite{Suleman2023} & 2023 & Smart MobiNet (MobileNet + Attention Mechanisms + Data Augmentation) & ISIC-2019 & 89\% accuracy \\
\cite{Riaz2023} & 2023 & CNN-LBP fusion model & HAM10000 & 97.42\% \\
\cite{Wang2023} & 2023 & CAM-based feature learning + History feature consistency learning & ISIC-2017, ISIC-2018 & 0.951 AUC \\
\cite{Vachmanus2023} & 2023 & DeepMetaForge: BEiT + DMFM metadata fusion & ISIC-2020, PAD-UFES-20, SKINL2, PH2 & 92\% F1-Score \\
\cite{Liu2023} & 2023 & ResNet-50 + Dual Attention + FA + DA + 2-layer ANN & 89 WSIs & 91.7\% accuracy \\
\cite{Hu2024} & 2024 & Multi-scale feature fusion with class-weighted loss and EfficientNetV2 & HAM10000 & 94.0\% accuracy \\
\cite{Tan2024} & 2024 & DGLA-ResNet50 (Dual-Branch Input + Global-Local Attention ResNet50) & ISIC-2018, ISIC-2019 & 90.71\% accuracy \\
\cite{Remya2024} & 2024 & Transfer learning framework based on ViT and channel attention & HAM10000 & 99\% accuracy \\
\cite{Capurro2024} & 2024 & Swin Transformer + Transfer Learning + Weighted Cross-Entropy Loss & Custom & 98.5\% accuracy \\
\cite{Suleiman2024} & 2024 & Two-step hierarchical binary classification model with transfer learning and RF & ISIC-2017 & 91.07\% accuracy \\
\cite{Du2025} & 2025 & RLFS + Dynamic data resampling + GCN-based feature selection & Custom & 98.37\% accuracy \\
\cite{Slian2025} & 2025 & CFPNet-M + Attention Gates + Dual-ACN Architecture & HFUS & 0.949 Entry Echo \\
\cite{Zhang2025} & 2025 & Multi-scale observation, intelligent focusing and feature screening & ISIC-2018, ISIC-2019 & 86.12\% accuracy \\
\cite{Li2025} & 2025 & Local-Global Fusion & ISIC-2018 & 93.10\% accuracy \\
\cite{Saeed2025} & 2025 & EfficientNetV2S, ResNet-50, Spatial-Context Group Attention & Custom & 98.20\% accuracy \\
\end{longtable}

As for multi-modality fusion, Vachmanus et al. proposed the DeepMetaForge, a novel vision transformer-based deep learning framework that fuses image data and patient metadata. In this study, the BEiT transformer is performed as an image encoder while the Deep Metadata Fusion Module (DMFM), which integrates metadata and image features through a compression and merging process, is used to simulate human perception of image-metadata interaction~\cite{Vachmanus2023}. Based on multiple imaging modalities data, Remya et al. introduced an innovative transfer learning framework for multimodal skin lesion analysis, incorporating ViTs, channel attention mechanisms, and region of interest (ROI) extraction. Where the ViTs leverage their self-attention mechanism to extract global image context. ROI extraction enhances lesion detection accuracy by focusing on the most relevant regions. Channel attention mechanisms improve feature selection by highlighting discriminative image characteristics~\cite{Remya2024}.

\subsection{Methods to solve class imbalance}\label{class_imbalance}

Class imbalance occurs when the number of samples in different classes of a dataset is highly uneven, leading to a bias in model training. This is a common issue in medical image classification, including skin lesion detection, where some diseases are significantly more frequent than others \cite{Ali2022}. The challenge has also been addressed by several studies mentioned above. For instance, the adaptive class-weighted loss function, which mitigates the effects of class imbalance by assigning higher penalties to underrepresented categories, is utilized to enhance classification accuracy in Qian’s study \cite{Qian2022}. Hu et al. additionally apply three strategies to address class imbalance. The first is class weighting, which assigns different loss weights to different classes to balance the model’s attention. Secondly, label smoothing is utilized to reduce overconfidence in the model’s predictions by distributing label probabilities. Finally, for adjusting training and test set distributions to mitigate data imbalance effects, resampling is also adapted \cite{Hu2024}. Table~\ref{tab:imbalance} lists the studies that focus on solving the class imbalance.

First of all, preprocessing technologies such as data augmentation, down-sampling, and over-sampling can be utilized to generate more samples for the minority class, thereby increasing its representation. Zunair and Ben Hamza generated synthetic images for the minority class using CycleGAN, which learns the mapping from benign to malignant lesions for unpaired image-to-image translation. A CNN guided by the focal loss function is then trained \cite{Zunair2020}. Qin et al. propose a skin lesion style-based GAN for data augmentation to address the challenges of skin lesion classification caused by scarcity of labeled data and class imbalance. The proposed approach integrates a style-based GAN for synthetic image generation with a transfer learning-based ResNet50 classifier to improve classification. The GAN model enhances data diversity and realism by generating high-resolution synthetic skin lesion images, addressing dataset scarcity and imbalance. These synthetic images are added to the training dataset to improve model generalization. The ResNet50 classifier learns robust feature representations from both real and synthetic images \cite{Qin2020}. Vasconcelos et al. propose a DCNN model based on GoogleNet, enhanced through data augmentation and transfer learning. Additionally, the introduction of a third class representing visual patterns outside the lesion area is applied to help address class imbalance, along with oversampling the minority class. Data augmentation techniques include geometric transformations, PCA-based color synthesis, and expert-guided image warping \cite{Vasconcelos2020}. Wu et al. proposed a GAN-based augmentation strategy that synthesizes realistic Raman spectroscopy spectra to expand the training set and improve classifier performance. Balanced augmentation is used to generate equal numbers of synthetic samples for each class, and stratified augmentation preserves the original class distribution. A CNN-based classifier is then trained using both real and synthetic samples, with comparisons against Logistic Regression (LR), Support Vector Machines (SVM), and LSTM-based models demonstrating up to 8.4\% improvement from GAN-based augmentation \cite{Wu2021}. For data preprocessing, Safdar et al. employed the Dull Razor algorithm for hair removal and median filtering for de-noising. K-means clustering is used to isolate the Region of Interest (ROI) by separating lesion from background via Euclidean distance-based clustering. For data augmentation, affine transformations, color space transformations, and contrast enhancement are used to address class imbalance and improve dataset diversity. For classification, a Majority Voting Ensemble combines the outputs of ResNet-50 and DenseNet-201 to distinguish benign and malignant lesions \cite{Safdar2021}. To address class imbalance and improve model generalization, Shen et al. proposed a low-cost, high-performance data augmentation strategy for deep learning-based skin lesion classification. Their framework introduced a two-stage augmentation search and model matching pipeline. The optimal augmentation strategy is determined via 5-fold cross-validation. In the network matching and fine-tuning stage, pre-trained EfficientNet models are refined using this optimal augmentation strategy \cite{Shen2022}. Due to the lack of dark skin images in dermatology resources, AI applications often underperform in diagnosing lesions in people of color. To address this, Rezk et al. developed two deep learning methods: Style Transfer (ST) and Deep Blending (DB). Using a VGG-19 network, ST extracts features from content and style images and iteratively blends them to create new images combining lesion content with dark skin style. DB integrates ST with Poisson image blending, which focuses on seamlessly merging lesion regions into dark skin backgrounds while minimizing texture changes \cite{Rezk2022}. To mitigate data limitations and enhance the generalization of a hybrid CNN system with six pretrained models, Nath et al. applied data augmentation with images of acne, blisters, cold sores, psoriasis, and vitiligo sourced online. Their system achieved 99\% accuracy on the validation set and 90\% on the test set, demonstrating strong performance across diverse conditions \cite{Nath2023}. In Raghavendra et al.'s work, the preprocessing pipeline included black-hat filtering to remove artifacts and data augmentation to mitigate class imbalance, increasing the training dataset to 45{,}756 images. A novel DCNN with optimized global average pooling was then constructed and trained on this expanded dataset \cite{Raghavendra2023}. Alwakid et al. proposed a deep learning pipeline integrating an Enhanced Super-Resolution Generative Adversarial Network (ESRGAN) for image quality enhancement and data augmentation. Two transfer learning-based models, Inception-V3 and InceptionResNet-V2, are used for prediction \cite{Alwakid2023}. Zhou et al. developed a Multi-Site Cross-Organ Calibrated Deep Learning (MuSClD) framework, using lung tissue from the test site to calibrate skin images from the training site. CycleGAN is employed to adjust visual properties such as color and brightness without introducing artifacts like blur or checkerboard patterns \cite{Zhou2023}. Akrout et al. proposed diffusion probabilistic models (DPMs) for synthetic image generation, enabling fine-grained control of the augmentation process using text prompts. Their pipeline first generates 30,000 synthetic images per disease category. A binary EfficientNet classifier filters out non-skin images, followed by an ensemble of EfficientNetV2, RegNet, and Swin-Transformer for further filtering. High-quality, correctly classified synthetic images are then added for data augmentation \cite{Akrout2024}. With psoriasis images far outnumbering lichen planus cases in the DermNet dataset, Eskandari et al. employed oversampling and undersampling to balance the dataset at 641 psoriasis and 378 lichen planus images. Data augmentation increased the data by 15\%, followed by ResNet-50 training for classification \cite{Eskandari2024}. Ju et al. presented a novel deep learning framework for early melanoma detection that combines dilated convolutions with Off-policy Proximal Policy Optimization (PPO) to address data imbalance challenges \cite{Ju2025}.  

Furthermore, the hybrid deep learning model is also effective in improving classification performance under the situation of imbalanced data. In Hoang’s research, ResNet50 and Atrous Convolutional Neural Network (ACNN) are integrated for disease candidate region extraction, forming a hybrid deep learning approach. In this architecture, ResNet50 is used to extract deep hierarchical features, while the ACNN refines segmentation with dilated convolutions for disease region enhancement. A fully connected network then processes extracted features for final classification. Additionally, common data augmentation including rotation, flipping, affine transformations, and color normalization is applied \cite{Hoang2020}. Xie et al. proposed a mutual bootstrapping deep convolutional neural network (MB-DCNN) model consisting of three key networks: the Coarse Segmentation Network (Coarse-SN) generates coarse lesion masks for initial localization; the Mask-Guided Classification Network (Mask-CN) uses Coarse-SN’s masks to refine lesion classification; and the Enhanced Segmentation Network (Enhanced-SN) integrates localization maps from Mask-CN to improve segmentation results. This study also introduced an optimized loss function combining Dice loss with rank loss to emphasize hard pixels \cite{Xie2020}. Based on a lightweight deep learning model, Sayed et al. chose SqueezeNet as the backbone and utilized the Bald Eagle Search (BES) optimization for hyperparameter tuning. Transfer learning is applied to enhance generalization by leveraging pre-trained SqueezeNet models. In data preprocessing, Random Over-Sampling and data augmentation are used to balance the dataset \cite{Sayed2021}. Al-Bander et al. proposed a benchmarking framework for evaluating deep learning algorithms for skin cancer detection using a hybrid multi-criteria decision-making (MCDM) approach that integrates entropy weighting and the VIKOR method, where cost-sensitive loss weighting and data augmentation are also involved to handle class imbalance \cite{AlBander2021}. Lafraxo et al. introduced the MelaNet framework combining CNN architecture with data augmentation to process imbalanced data, including geometric and color augmentation techniques \cite{Lafraxo2022}. Alam et al. established a hybrid deep learning approach with data augmentation to enhance feature learning and improve skin cancer classification performance. The hybrid model combines CNN architectures to extract robust features from lesion images, then uses fine-tuned feature selection mechanisms to handle intra-class variation. Additionally, data augmentation and weighted loss functions are deployed to reduce bias toward majority classes \cite{Alam2022}. Nie et al. constructed a hybrid CNN-Transformer model integrating Focal Loss (FL), achieving a 15\% accuracy improvement over the baseline model. In this approach, a CNN and ViT extract local features and global contextual information from dermoscopic images respectively, with a Multi-Layer Perceptron (MLP) classifier employed at the final prediction stage. The FL loss function reduces model bias by dynamically adjusting the loss contribution of easy and hard samples \cite{Nie2023}. Mahmud et al. compared the performance of VGG16 and ResNet50 for dermoscopic image classification on the HAM10000 dataset, implementing random resampling and optimizing the models through transfer learning and fine-tuning \cite{AlMahmud2024}. Hussain et al. leveraged DenseNet, MobileNet, NASNet, and EfficientNet in their hybrid model, also optimizing data augmentation and hyperparameter tuning \cite{Hussain2025}.

Moreover, several studies have modified the loss function or applied class weights to adjust the learning process to focus on minority classes. Gouabou et al. integrated both data-level and classifier-level techniques to address imbalanced data. Specifically, at the data level, they utilized oversampling methods such as SMOTE and advanced augmentation strategies like Mixup and Cutmix to synthetically balance the dataset. At the classifier level, they applied cost-sensitive learning, focal loss, and label-distribution-aware margin loss (LDAM) to increase sensitivity toward minority classes \cite{Gouabou2022}. Gayatri and Aarthy addressed the challenge of overfitting in the highly imbalanced ISIC-2019 dataset for skin cancer detection using ResNet-50 combined with focal loss and data augmentation \cite{Gayatri2024}. Askale et al. combined SKDCNET, FVGG16, and InceptionV3, employing a hard voting scheme to aggregate predictions for more robust results. Specifically, the three CNNs are trained independently, and their outputs are combined using a majority voting mechanism for final classification. In addition, median filter noise removal and data augmentation are used to enhance preprocessing \cite{Askale2024}. Selvaraj et al. presented a comprehensive deep learning ensemble framework consisting of six CNNs, including both average and weighted ensemble models optimized via grid search. Cost-sensitive learning and data augmentation are also employed to address imbalanced data, along with hair removal using blackhat filtering and image inpainting to reduce noise \cite{Selvaraj2024}. Akinrinade and Du utilized Few-Shot Learning (FSL) to improve generalization with limited training data, integrating it with data augmentation techniques. GANs are also applied to generate samples for underrepresented classes. After this preprocessing, a CNN architecture is used to extract texture and lesion-based features for classification. Transfer learning with pre-trained models is employed for feature extraction and fine-tuning, along with oversampling and loss function adjustments \cite{Akinrinade2025}.

\begin{longtable}{%
>{\centering\arraybackslash}m{1.5cm}   
>{\centering\arraybackslash}m{1.3cm} 
>{\centering\arraybackslash}m{7cm}   
>{\centering\arraybackslash}m{3cm}   
>{\centering\arraybackslash}m{2.5cm}}  
\caption{Included studies mainly focusing on solving class imbalance.}
\label{tab:imbalance} \\
\toprule
\textbf{Reference} & \textbf{Year} & \textbf{Method} & \textbf{Datasets} & \textbf{Results} \\
\midrule
\endfirsthead

\multicolumn{5}{c}{{\bfseries Table \thetable\ (continued)}} \\
\toprule
\textbf{Reference} & \textbf{Year} & \textbf{Method} & \textbf{Datasets} & \textbf{Results} \\
\midrule
\endhead

\bottomrule
\endfoot

\cite{Hoang2020} & 2020 & Hybrid deep learning (ResNet50 + ACNN) with data augmentation for segmentation and disease region extraction & ISIC-2018 & 0.9996 Jaccard Index \\
\cite{Zunair2020} & 2020 & CycleGAN + CNN & ISIC-2016 & 0.8118 AUC \\
\cite{Qin2020} & 2020 & Skin Lesion Style-Based GAN + Transfer Learning with ResNet50 & ISIC-2018 & 95.2\% accuracy \\
\cite{Xie2020} & 2020 & MB-DCNN & ISIC-2017, PH2 & 0.977 AUC \\
\cite{Vasconcelos2020} & 2020 & Ensemble CNN + Transfer Learning + Data Augmentation + Oversampling + Third Class introduction & ISBI-2016 & 0.825 accuracy \\
\cite{Sayed2021} & 2021 & SqueezeNet + BES + Over-Sampling & ISIC-2020 & 98.37\% accuracy \\
\cite{Wu2021} & 2021 & GAN-based data augmentation + CNN for Raman Spectroscopy skin cancer classification & RS data from Mohs Surgery Center & 0.945 AUC \\
\cite{Safdar2021} & 2021 & Dull Razor + Data Augmentation + Majority Voting Ensemble & PH2 & 95.2\% accuracy \\
\cite{AlBander2021} & 2021 & Benchmarking 12 CNNs using Entropy + VIKOR (MCDM) & HAM10000 & InceptionResNetV2 ranked 1 \\
\cite{Lafraxo2022} & 2022 & MelaNet & ISIC-2017, PH2, MED-NODE & 98.44\% accuracy \\
\cite{Alam2022} & 2022 & Hybrid Deep Learning (CNN-based approach) with Data Augmentation for class balancing & HAM10000 & 91\% accuracy \\
\cite{Shen2022} & 2022 & Low-Cost-Augment (LCA) strategy + EfficientNet-based deep learning for skin lesion classification & HAM10000, ISIC-2017, Derm7pt & 0.909 AUC \\
\cite{Gouabou2022} & 2022 & CNN + SMOTE + Mixup/CutMix + Cost-sensitive Loss + LDAM & ISIC-2018 & 87\% accuracy, 2\% superior to baselines \\
\cite{Bozkurt2023} & 2023 & Inception-ResNet-V2 with affine transformation-based data augmentation & HAM10000 & 95.09\% accuracy, 11.5\% improvement \\
\cite{Rezk2022} & 2023 & Style Transfer (ST) and Deep Blending (DB) & DermNet NZ, ISIC-2018, Dermatology Atlas & 76\% accuracy \\
\cite{Nath2023} & 2023 & CNN-based system using 6 pretrained models + data augmentation & Custom & 99\% accuracy \\
\cite{Raghavendra2023} & 2023 & DCNN with global average pooling + black hat filtering + data augmentation & HAM10000 & 97.2\% accuracy \\
\cite{Nie2023} & 2023 & Hybrid CNN-Transformer model with Focal Loss for class imbalance mitigation & ISIC-2018 & 89.48\% accuracy (baseline 74.21\%) \\
\cite{Alwakid2023} & 2023 & ESRGAN + Inception-V3 + InceptionResNet-V2 & HAM10000 & 91.26\% accuracy \\
\cite{Zhou2023} & 2023 & MuSClD + CycleGAN & Custom & 0.92 AUC \\
\cite{Gayatri2024} & 2024 & ResNet-50 + Data Augmentation + Focal Loss & ISIC-2019 & 98.85\% accuracy \\
\cite{Askale2024} & 2024 & SKDCNET, FVGG16, InceptionV3 Ensemble & Custom & 98\% accuracy \\
\cite{AlMahmud2024} & 2024 & Transfer learning and fine-tuning of VGG16 \& ResNet50 with random resampling for class balancing & HAM10000 & 94.69\% accuracy \\
\cite{Akrout2024} & 2024 & Diffusion-based data augmentation + Deep ensemble classification & Custom & 95\% accuracy more \\
\cite{Eskandari2024} & 2024 & ResNet-50 + oversampling + undersampling & DermNet & 89.07\% accuracy \\
\cite{Selvaraj2024} & 2024 & Ensemble CNN & HAM10000 & 0.97 AUC \\
\cite{Akinrinade2025} & 2025 & CNN-based deep learning model with transfer learning, data augmentation, GANs, and few-shot learning & ISIC-2017, HAM10000 & 0.956 AUC \\
\cite{Ju2025} & 2025 & Dilated convolutions, Off-policy PPO & ISIC-2020 & 91.84\% F-measure \\
\cite{Hussain2025} & 2025 & DenseNet, MobileNet, NASNet, and EfficientNet & ISIC-2020 & 99.59\% accuracy \\
\end{longtable}

\section{Prospective techniques}\label{Prospective techniques}

\begin{figure}[htbp]
\centering
\includegraphics[width=0.8\textwidth]{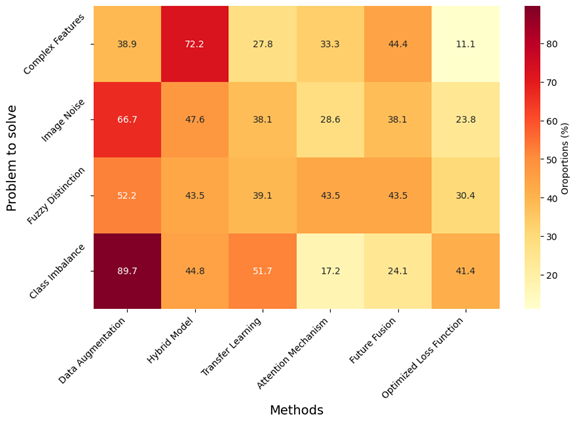}
\caption{Heat map of the application of prospective techniques in dermatologic diagnostics.}\label{fig:technology_distribution}
\end{figure}

In these 91 articles, 18 of them are focused on dealing with the complex features, while 21 articles are coping with the image noise. The number of articles to solve the intra-class variation and inter-class similarity is 23. It is obvious that class imbalance is one of the most attractive challenges to solve, with 29 articles listed above. In the reviewed articles, we found that \textit{data augmentation}, \textit{hybrid model}, \textit{transfer learning}, \textit{attention mechanism}, \textit{feature fusion}, and \textit{optimized loss function} are the most widely deployed technologies to solve the four challenges aforementioned. Fig.~\ref{fig:technology_distribution} demonstrates the application of various technologies in the solutions of these four different problems.

\subsection{Data augmentation}

\begin{figure}[htbp]
\centering
\includegraphics[width=1\textwidth]{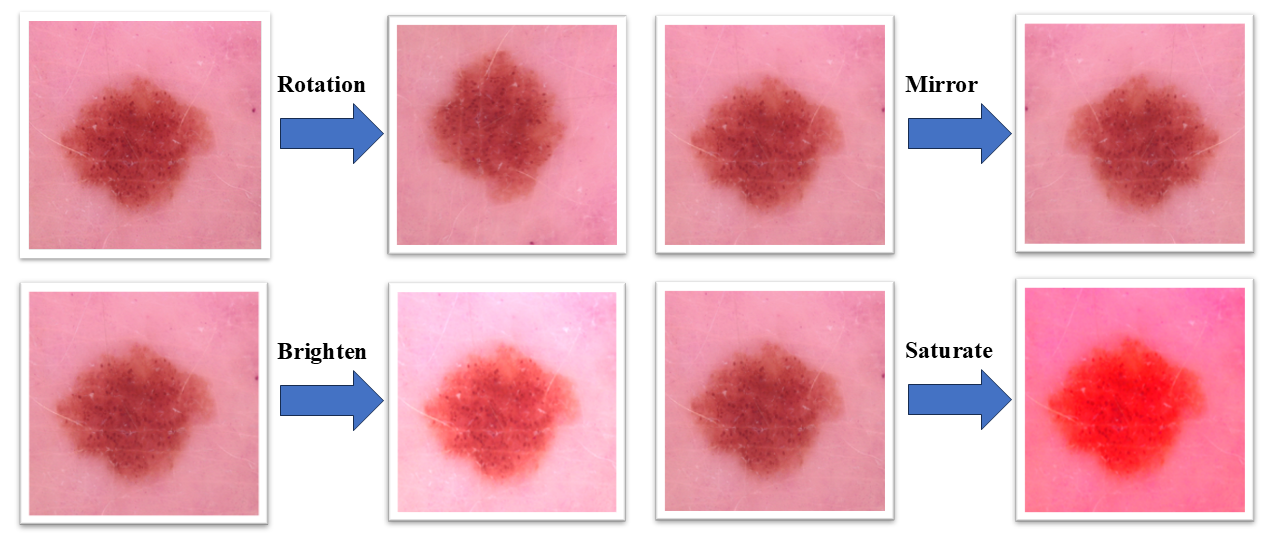}
\caption{Example of data augmentation.}\label{fig:augmentation_example}
\end{figure}

Given the 91 studies retrieved above, 59 papers (72.84\%) utilized data augmentation, and 89.7\% of the reviewed papers addressing class imbalance adopted this technique. This underscores its importance in deep learning-based dermatological diagnosis. Fig.~\ref{fig:augmentation_example} shows an example of data augmentation. Data augmentation refers to the manual extension of a dataset by applying transformations to existing data. Future research should focus on improving augmentation strategies to better handle rare disease classes and diverse clinical scenarios. 

Recent works have explored advanced generative approaches, including Generative Adversarial Networks (GANs) and diffusion models, which can synthesize realistic and diverse skin lesion images \cite{Zhang2022}. Integrating these models with clinical priors may produce more representative samples, enhancing fairness across skin types and mitigating data bias. 

Moreover, task-adaptive augmentation policies guided by reinforcement learning or meta-learning can dynamically optimize transformation parameters for specific diagnostic objectives. Combining such advanced augmentation methods with federated or privacy-preserving learning frameworks may further enable collaborative dataset enrichment without breaching patient confidentiality. These directions hold promise for improving model robustness, generalization, and real-world applicability in automated dermatological diagnosis \cite{Feng2024}.

\subsection{Hybrid model}

With the growing complexity of diagnostic challenges, 46 of the 91 reviewed studies (50.54\%) employed hybrid models, and 72.2\% of the papers addressing complex feature problems adopted this approach. This highlights the increasing importance of hybrid architectures in enhancing dermatological diagnosis performance. 

Future research should focus on advancing hybrid modeling beyond simply combining existing architectures such as Convolutional Neural Networks (CNNs), Vision Transformers (ViTs), and Recurrent Neural Networks (RNNs). Promising directions include dynamically adaptive hybrid networks that can automatically select or weight model components based on lesion characteristics or patient demographics \cite{Li2025}. Additionally, integrating hybrid models with multi-scale feature fusion and attention mechanisms may further improve the ability to capture both local lesion details and long-range contextual information, enhancing diagnostic reliability \cite{Nerella2024}. Another prospective avenue is designing lightweight high-performing hybrid models, which is suitable for edge devices, enabling real-time clinical applications. Combining hybrid networks with explainable AI techniques will also be essential to improve interpretability and foster trust in automated diagnosis. These advancements could significantly enhance robustness against intra-class variation, inter-class similarity, and other complex data challenges, driving more clinically applicable deep learning solutions for dermatology.

\subsection{Transfer learning}

In this review, 37 studies employed transfer learning techniques, all reporting strong performance. Among them, 51.7\% of the 29 studies addressing class imbalance and 39.1\% of the 23 studies tackling fuzzy distinction (intra-class variation and inter-class similarity) leveraged transfer learning. This demonstrates its significant role in enhancing diagnostic accuracy and generalization in dermatological AI applications. 

Future work should move beyond conventional fine-tuning of large-scale pre-trained models \cite{Bhatt2023}. Promising directions include domain-adaptive transfer learning that incorporates dermatology-specific priors to reduce negative transfer effects, and self-supervised or few-shot transfer methods to improve performance with minimal labeled data. Cross-modal transfer learning—transferring knowledge between dermoscopic images, clinical photographs, and histopathology slides—may also improve robustness for multi-task diagnostic pipelines. Furthermore, developing lightweight transfer learning strategies optimized for deployment on edge devices would support real-time, point-of-care diagnosis. Combining transfer learning with explainable AI techniques could enhance model transparency, making it easier for clinicians to interpret transferred features and trust automated diagnostic results. These advancements are expected to further improve scalability and clinical adoption of deep learning in dermatology.

Moreover, when applying transfer learning to new datasets or imaging conditions, a key challenge is to ensure cross-dataset generalization under domain shift to maintain robust performance across varying data distributions. In recent years, the challenge of cross-dataset generalization and domain shift in dermatological image analysis has become increasingly prominent. For example, a new dataset called IMPS has been created by combining diverse clinical image sources via a new dataset. After applying Domain‑Adversarial Neural Network, an unsupervised domain adaption method, the generalisation could be enhanced \cite{Sultana}. In addition, the DRMv2Net model also demonstrates strong cross-dataset generalization. This model is evaluated on both the ISIC 2357 and PAD-UFES 20 datasets and achieving accuracies of 96.11\% and 96.17\% respectively, thereby indicating robustness against domain shift across different imaging sources. \cite{Bhargavi2025}. Prompt-based methods built on global and local prompts can effectively separate domain-shared and domain-specific knowledge. They achieve superior segmentation generalization across five heterogeneous datasets \cite{Zhao2025PromptDG}. Moreover, Fourier-domain style transfer \cite{Wang2023CFDA} and structure-preserving image translation \cite{Ilyas2024UDA} further enhance model adaptability and robustness under varying imaging protocols, devices, and magnification levels \cite{Kang2023SPIT}.

In a word, the convergence of transfer learning and cross-dataset generalization presents a promising frontier for dermatological AI. Transfer learning enables the reuse of pretrained representations and accelerates model development with limited labeled data, while cross-dataset generalization addresses the critical problem of domain shift when models encounter new imaging sources or populations.

\subsection{Attention mechanism}

In this review, 27 studies employed attention mechanisms, with 10 of them specifically targeting challenges of intra-class variation and inter-class similarity. This highlights the attention mechanism’s effectiveness in improving model focus and diagnostic performance in dermatological analysis. 

Future research should move beyond basic spatial or channel attention modules \cite{Cai2024,R. Liu}. Advanced approaches such as multi-scale hierarchical attention, transformer-based self-attention, and cross-modal attention could be explored to better capture complex lesion patterns and integrate information from dermoscopic, clinical, and histopathological images. Additionally, adaptive attention mechanisms that dynamically adjust their focus based on lesion type, image quality, or patient demographics may further enhance model robustness and fairness. Combining attention with explainable AI techniques could improve interpretability and clinical trust. Exploring lightweight attention designs optimized for real-time applications on portable medical devices is another promising direction. These advancements are expected to make attention-driven models more scalable and clinically relevant in automated dermatological diagnosis.

In addition to the studies originally reviewed, we further examined recent advances that explicitly integrate hierarchical attention with multi-scale feature design, which aligns closely with the reviewers’ suggestions. For example, AMST-Net introduces a dual-encoder architecture in which a multi-scale Transformer generates hierarchical attention features that are progressively fused with CNN representations, demonstrating the effectiveness of combining layered self-attention with multi-resolution information for improving boundary precision and global context modeling. These complementary findings reinforce the growing trend toward adaptive, multi-level attention frameworks in dermatological imaging and support our expanded discussion on attention hierarchies and multi-scale designs in automated skin lesion analysis \cite{Gao2026}. The MTA-Net model explicitly integrates multi-scale triplet spatial and channel attention, enabling the network to capture fine-grained lesion characteristics across different scales and dimensions. By enriching both spatial and channel-wise representations, MTA-Net significantly improves multiclass skin lesion classification performance on HAM10000 and ISIC2019, demonstrating the effectiveness of hierarchical multi-scale attention in handling high inter-class similarity and intra-class variability \cite{Gajera2025}. HMSA-Net introduces a hierarchical multi-scale aggregation framework that integrates Res2Net-based multi-scale contextual encoding with Hierarchical Attention Refinement Blocks (HARBs), enabling dynamic channel recalibration through squeeze-and-excitation mechanisms. By fusing features across multiple resolutions and emphasizing anatomically critical structures, the model demonstrates strong segmentation performance across diverse biomedical imaging modalities, including ISIC-2018 skin lesion segmentation \cite{Magdy2026}. There are also several studies proves that multi-scale attention mechanisms, including structures such as multi-scale spatial pyramid attention \cite{Yu2024MSPAM,Qian2024SPCBNet} and hierarchical attention modules \cite{Huang2025MADenseUNet}, effectively fuse features across scales and levels. They enhance the model's ability to recognize complex skin lesions. By capturing both local details and global context, these mechanisms improve detection of small lesions and blurred boundaries, and they strengthen the model’s generalization across diverse disease types.

Overall, these recent developments illustrate that combining hierarchical attention with multi-scale feature aggregation provides a powerful mechanism for capturing both global contextual cues and subtle local lesion details. This convergence of attention-driven and multi-scale design principles is becoming a key direction for building more robust, generalizable, and clinically reliable dermatological deep learning models.




\subsection{Feature fusion}

In this review, 44.4\% of the studies addressing complex features employed feature fusion techniques, highlighting its importance in improving diagnostic accuracy for dermatological applications. Rather than focusing solely on integrating multiple feature types, future research should explore more advanced and adaptive fusion strategies \cite{Saba2021}. 

Promising directions include dynamic feature fusion frameworks that can automatically weigh local and global features based on lesion characteristics, as well as cross-modal fusion methods that integrate dermoscopic images, metadata, and clinical records in a task-aware manner. Multi-scale and hierarchical fusion approaches, potentially combined with attention mechanisms or graph neural networks, could further enhance the model’s ability to capture complex dermatological patterns and improve interpretability.



\subsection{Optimized loss function}

In this review, 41.4\% of the papers addressing class imbalance incorporated optimized loss functions, underscoring their importance in guiding model learning for dermatological diagnosis. Rather than focusing on conventional loss formulations, future research should explore advanced, adaptive loss designs tailored to the unique challenges of skin lesion analysis \cite{Saba2021}. 

Promising directions include dynamic loss weighting mechanisms that automatically adjust penalties based on sample difficulty, lesion rarity, or inter-class similarity during training. Combining multiple loss objectives—such as focal, Dice, contrastive, and margin-based losses—within multi-task frameworks could further improve classification and segmentation accuracy for rare dermatological conditions. For instance, dice-based loss functions are also widely adopted to mitigate class imbalance, especially in lesion segmentation tasks where the foreground region is small relative to the background. In addition, integrating uncertainty-aware or noise-robust loss functions can enhance resilience to mislabeled or low-quality data, while fairness-driven loss formulations may help mitigate bias across different skin tones and demographics. Designing computationally efficient, task-specific loss functions suitable for federated and real-time applications will also be crucial for broader clinical adoption.




\section{Frontier Advances and Emerging Paradigms}\label{Frontier Advances and Emerging Paradigms}

The rapid evolution of dermatological artificial intelligence in recent years has given rise to a new generation of frontier technologies that extend far beyond conventional deep learning solutions. These emerging paradigms integrate multi-dataset fusion, cross-domain generalization, attention-driven hybrid architectures, and uncertainty-aware learning, collectively reshaping the landscape of automated skin disease diagnosis. Unlike earlier methods that rely on single-source data or isolated network designs, current advances emphasize robustness, adaptability, and clinical reliability through the integration of diverse feature sources, hierarchical attention mechanisms, and domain-shift-resistant modeling strategies. This section highlights these transformative developments, outlining how they mark a paradigm shift toward more comprehensive, scalable, and clinically deployable diagnostic systems.

\subsection{Multimodal and Cross-Domain Fusion}
Recent work has shown that multi-dataset fusion and cross-domain learning can substantially enhance the robustness of dermatological deep learning models. A representative example is EffiCAT (EfficientNet Concatenation Attention Technology), a hybrid architecture designed to improve generalization across heterogeneous datasets. EffiCAT integrates complementary feature representations extracted from EfficientNet-B0 and EfficientNet-B4 through feature concatenation, followed by multiple attention modules including a dual channel attention layer and CBAM to refine salient lesion characteristics. EffiCAT demonstrates strong performance, achieving 94.48\% accuracy on the merged dataset and maintaining high cross-dataset robustness with 92.08\% accuracy on ISIC-2018 \cite{EffiCAT2025}. Recent studies beyond EffiCAT have further demonstrated the value of multimodal and cross-domain fusion in dermatological AI. Several works jointly model dermoscopic images, clinical photographs, and patient metadata through attention-based or multi-stage fusion strategies, such as adversarial attention fusion networks, tri-modal cross-attention transformers (TMCT), and hierarchical fusion architectures like MsfuseNet \cite{Msfusenet}. Cai et al. also built a hybrid skin cancer segmentation model that integrates CNNs, Transformers, and a U-Net framework with a dual-frequency cross-fusion strategy.By combining features extracted from the Discrete Walsh Transform and the Haar wavelet transform, the model leverages both high-frequency details and low-frequency structural information, while an attention mechanism enhances the fusion of these complementary features \cite{Cai2026}. Recent studies also show that fusing skin-disease images with patient metadata can markedly improve accuracy and robustness. In specific, cross-attention mechanisms \cite{Yu2025Deep}, mutual-information switching modules \cite{TranVan2025}, transformer architectures \cite{Verma2025}, enable dynamic integration of heterogeneous features and capture richer lesion information. Some models now achieve over 95\% accuracy on public datasets such as HAM10000 and ISIC 2018, with reports approaching 99

Multimodal and cross-domain fusion carries substantial technical importance in advancing dermatological artificial intelligence \cite{xie2025chat}. By integrating heterogeneous information sources such as dermoscopic images, clinical photographs, and patient-level metadata, fusion-based models capture complementary feature representations that cannot be derived from any single modality alone. This integrated learning process improves resilience to domain shifts. Meanwhile, dataset-specific bias, and enhances generalization across diverse clinical environments can also be reduced by this technology. Moreover, cross-domain alignment enables models to learn invariant patterns shared across imaging conditions, acquisition devices, and patient populations, thereby strengthening diagnostic reliability.

\subsection{Uncertainty-aware and Reliability-calibrated Learning}
Although interpretability has gained increasing attention, recent studies emphasize that reliability assessment. For instance, uncertainty estimation in deep-learning models solving skin lesion recognition enables the prediction not only of labels but also of the model’s confidence, distinguishing between data-driven uncertainty and model-driven uncertainty \cite{Asgharnezhad2025},\cite{Zoravar2025}.

Beyond estimating uncertainty, it is vital to ensure that confidence scores are calibrated. For example, in A novel uncertainty‑aware deep learning technique with an application on skin cancer diagnosis the authors develop a Monte Carlo Dropout (MCD) based framework augmented with an uncertainty-aware loss which seeks to reduce expected calibration error (ECE) and improve uncertainty accuracy (UA) \cite{Shamsi2025}. These findings underscore that for clinical deployment, systems must not only achieve high accuracy, but also provide reliable, calibrated confidence estimates to support risk-aware decision making. By enabling reliable confidence assessment, uncertainty-aware models help identify cases where predictions may be untrustworthy and thus trigger expert review, which is essential for safe clinical deployment of AI systems.

Besides, the combination of multimodal fusion and diagnostic uncertainty quantification further improves accuracy, reliability, and clinical utility. This integration provides a solid foundation for the safe deployment of intelligent dermatological diagnosis \cite{Zuo2025,Sikha2025}.

\section{Discussion and Conclusion}\label{Discussion and Conclusion}
\subsection{Historical progression}
\begin{figure}[htbp]
\centering
\includegraphics[width=1\textwidth]{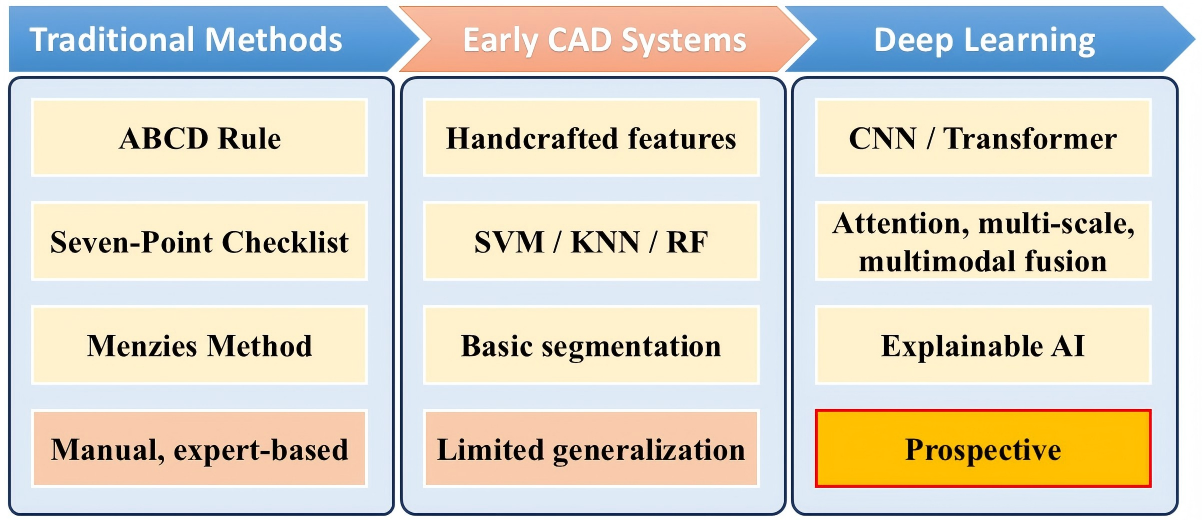}
\caption{Progression from traditional diagnostic techniques to deep learning.}\label{fig:developing_progression}
\end{figure}

The Fig.\ref{fig:developing_progression} illustrates the methodological evolution from rule-based clinical assessment to early computer-aided diagnosis using handcrafted features, and finally to modern deep learning and explainable AI models that integrate attention mechanisms, multimodal fusion, and interpretability tools such as Grad-CAM and SHAP. 

This historical progression highlights that performance gains alone are not sufficient for clinically reliable skin disease diagnosis. Traditional rule-based methods offered transparency but lacked robustness, whereas early CAD systems and deep learning models improved accuracy at the cost of increased opacity. Recent advances in explainable and uncertainty-aware AI suggest that future work must balance accuracy with interpretability, calibration, and cross-domain robustness so that automated systems can truly support dermatologists in safe, trustworthy clinical decision-making.

\subsection{Technological summary}

Now, the application of deep learning in dermatology has shown significant promise in improving the accuracy and efficiency of skin disease diagnosis. However, several challenges remain that must be addressed to further enhance clinical applicability and reliability. This review systematically analyzed 91 articles published between 2020 and mid-2025, identifying key obstacles including complex features, image noise, intra-class variation, inter-class similarity, and class imbalance.  

\begin{table}[htbp]
\centering
\caption{Distribution of studies across four key challenges in automated skin disease diagnosis.}
\label{Distribution_of_studies_4_challenges}
\begin{tabular}{@{}cc@{}}
\toprule
\textbf{Research Theme} & \textbf{Number of Studies} \\
\midrule
Complex Features & 18 \\
Image Noise & 21 \\
Intra-Class Variation and Inter-Class Similarity & 23 \\
Class Imbalance & 29 \\
\bottomrule
\end{tabular}
\end{table}

The Table \ref{Distribution_of_studies_4_challenges} presents the distribution of 91 studies across four major challenges in automated skin disease diagnosis: complex features, image noise, intra-class variation, and inter-class similarity. Notably, the largest number of studies (29) addresses class imbalance, indicating that this issue is a primary concern in the field. A significant portion of the studies (23) also focuses on intra-class variation and inter-class similarity, highlighting the difficulty in distinguishing lesions within the same class. Image noise is another common challenge, with 21 studies dedicated to mitigating it. In contrast, complex features received comparatively less attention, with only 18 studies focusing on this area, suggesting that it remains a more complex and perhaps less explored issue in dermatological deep learning applications.

Techniques such as data augmentation, hybrid models, transfer learning, attention mechanisms, feature fusion, and optimized loss functions have demonstrated effectiveness in overcoming these challenges. In particular, data augmentation has proven crucial for addressing class imbalance and generating high-quality synthetic images for underrepresented skin conditions. Future work should explore more sophisticated augmentation methods—such as GANs, diffusion models, and multimodal data synthesis that integrates dermoscopic images, metadata, genetic information, and histopathology—to build richer and more informative datasets.  

Capturing both local and global lesion features remains critical to improving diagnostic performance \cite{Xu2023}. Advanced hybrid architectures and ensemble learning methods, including optimized transfer learning strategies, can further enhance model robustness. Attention mechanisms and feature fusion approaches should be refined to dynamically adapt to lesion characteristics and multimodal data sources, improving both interpretability and diagnostic accuracy. Moreover, Explainable AI (XAI) techniques \cite{Ieracitano2025} should be integrated into future diagnostic systems to make model predictions transparent for clinicians, thereby fostering trust and promoting real-world adoption.  

In summary, deep learning has significantly transformed dermatological diagnostics, offering advances in accuracy, efficiency, and scalability. Recent innovations in Transformer-based architectures, optimization-driven CNNs, and multimodal hybrid frameworks have achieved state-of-the-art performance across multiple benchmark datasets, demonstrating strong potential for clinical translation. Furthermore, developments in data augmentation, feature fusion, and interpretability have contributed to the robustness of automated diagnostic systems.  

Despite these achievements, important challenges remain, particularly in improving model generalization across diverse skin tones, adapting to real-world clinical variability, and addressing data scarcity for rare disease classes. Future research should prioritize enhancing model fairness, leveraging cross-domain knowledge, developing lightweight models for real-time applications, and enabling privacy-preserving collaborative training across institutions. With continued interdisciplinary collaboration and technical innovation, deep learning is poised to play an increasingly pivotal role in enabling early detection, personalized treatment, and improved patient outcomes in dermatological care.

\subsection{Comments}
Methods addressing complex features benefit from hybrid CNN-Transformer architectures, multi-scale feature fusion, and attention-based representations, which collectively enhance the extraction of irregular lesion shapes, color variations, and global contextual cues. These models achieve strong performance but often require high computational resources and large datasets. Their interpretability mainly relies on attention maps or CAM-based visualizations, which provide partial but not fully transparent insights into decision-making.

Image noise is targeted by technologies including preprocessing pipelines, contrast enhancement, denoising filters, and knowledge-distillation-based light-weighted models-effectively improve robustness to artifacts such as hair, illumination variations, and background clutter. However, they frequently depend on handcrafted preprocessing steps, which may limit generalizability across devices or clinical environments. Interpretability remains limited because many enhancements occur at the pixel or feature level, which is less intuitive for clinicians.

Approaches for intra-class variation and inter-class similarity utilize attention mechanisms, multi-scale learning, metadata fusion, and hierarchical classification strategies to better capture subtle local structures and reduce ambiguity between visually similar lesion types. These designs are clinically relevant, as they mirror the multi-scale visual reasoning used by dermatologists. However, the increased architectural complexity may reduce model transparency. Attention-based explanations enhance interpretability, but they do not necessarily ensure causal clinical correctness.

Methods addressing class imbalance rely on data augmentation, GAN or diffusion-based synthesis, weighted loss functions, and resampling strategies. Such techniques substantially improve minority-class sensitivity and prevent models from being biased toward common lesion types. Yet, synthetic samples may introduce distributional shifts or unrealistic artifacts, and weighted losses can over-correct the imbalance. Interpretability is generally not directly affected, but clinical reliability requires verification to ensure that augmented images do not mislead the model.

Overall, as listed in Table~\ref{tab:challenge_matrix}, recent advances show strong technical performance, their practical and clinical reliability depends on consistent interpretability, robustness across diverse patient groups, and validation under real-world conditions. These considerations remain essential for translating deep learning systems into trustworthy dermatological decision support tools.

\begin{sidewaystable}[h]
    \centering
    \caption{Compact comparative matrix of deep-learning methods addressing 4 major challenges.}
    \label{tab:challenge_matrix}
    \small

    \begin{tabular}{%
        >{\centering\arraybackslash}m{0.18\textwidth} 
        >{\centering\arraybackslash}m{0.20\textwidth} 
        >{\centering\arraybackslash}m{0.18\textwidth} 
        >{\centering\arraybackslash}m{0.15\textwidth} 
        >{\centering\arraybackslash}m{0.15\textwidth} 
        >{\centering\arraybackslash}m{0.15\textwidth} 
    }
        \toprule
        \textbf{Challenge} &
        \textbf{Representative Techniques} &
        \textbf{Performance} &
        \textbf{Robustness} &
        \textbf{Interpretability} &
        \textbf{Clinical Reliability} \\
        \midrule
        \textbf{Complex Feature Representation} &
        CNN--Transformer hybrids; multi-scale fusion; attention mechanisms &
        \textbf{Strong}: typically higher accuracy than CNN baselines &
        \textbf{Moderate} &
        \textbf{Weak} &
        \textbf{Moderate} \\
        \addlinespace[2pt]
        \textbf{Noise \& Artifact Robustness} &
        Preprocessing pipelines; contrast enhancement; denoising; knowledge distillation &
        \textbf{Moderate}: mainly improves performance under noisy inputs &
        \textbf{Strong} &
        \textbf{Weak} &
        \textbf{Weak} \\
        \addlinespace[2pt]
        \textbf{Fine-grained Visual Discrimination} (Intra-/Inter-class variation) &
        Multi-scale learning; hierarchical attention; metadata fusion &
        \textbf{Strong}: better separation of visually similar lesions &
        \textbf{Moderate} &
        \textbf{Weak} &
        \textbf{Moderate} \\
        \addlinespace[2pt]
        \textbf{Class Imbalance} &
        Data augmentation; GAN/diffusion synthesis; weighted loss; resampling &
        \textbf{Moderate}: higher minority-class sensitivity; overall metrics vary &
        \textbf{Moderate} &
        \textbf{Moderate} &
        \textbf{Weak} \\
        \bottomrule
    \end{tabular}

\end{sidewaystable}

\subsection{Conclusion}
Overall, this study makes a distinctive contribution by establishing a comprehensive, PRISMA-based challenge–solution taxonomy for deep learning in dermatology, bridging technical innovation with clinical applicability. This framework may serve as a foundation for future research toward reliable, interpretable, and equitable dermatological AI systems.

\section*{Acknowledgement}
Support was provided by National Key Research and Development Program of China[http://dx.doi.org/10.13039/501100013290] (Grant No. 2024YFC2511003) and Natural Science Foundation of Zhejiang Province (Project No. LDT23E05015A01). The authors declare that there are no conflicts of interest or competing financial interests related to this study. The authors gratefully acknowledge the constructive comments and valuable suggestions provided by the anonymous reviewers, associate editors, and the editor-in-chief, which have significantly contributed to improving the quality of this manuscript.

This work was previously available as a preprint on arXiv. 

https://arxiv.org/abs/2510.03869

\appendix







\section*{Appendix}

\begin{longtable}{|>{\raggedright\arraybackslash}p{7cm}|>{\raggedright\arraybackslash}p{7cm}|}
\hline
\textbf{Checklist Item} & \textbf{Location in Paper} \\
\hline
\endfirsthead
\hline
\textbf{Checklist Item} & \textbf{Location in Paper} \\
\hline
\endhead
\hline
\endfoot
\hline

\textbf{Title} & \\
Identify the report as a systematic review. & The title clearly states that it is a systematic review of deep learning in automated skin disease diagnosis. \\
\hline
\textbf{Abstract} & \\
See the PRISMA 2020 for Abstracts checklist. & The abstract includes background, objective, methods, results, and conclusion. \\
\hline
\textbf{Introduction} & \\
Describe the rationale for the review in the context of existing knowledge. & The introduction describes the challenges in automated skin disease diagnosis and why this review is needed. \\
\hline
Provide an explicit statement of the objective(s) or question(s) the review addresses. & The objective of the review is stated as evaluating deep learning approaches for automated skin disease diagnosis. \\
\hline
\textbf{Methods} & \\
Specify the inclusion and exclusion criteria for the review. & The eligibility criteria for included studies are detailed, including the year range (2020–2025) and the focus on deep learning methods for skin disease diagnosis. \\
\hline
Specify all databases, registers, websites, organisations, reference lists, and other sources searched or consulted to identify studies. & Databases such as PubMed, IEEE Xplore, and Web of Science are listed with the date of last search. \\
\hline
Present the full search strategies for all databases. & The full search strategies for the included databases are detailed. \\
\hline
Specify the methods used to decide whether a study met the inclusion criteria. & The study selection process is explained, with a PRISMA flow diagram included in Section~\ref{Methods}. \\
\hline
Specify the methods used to collect data from reports. & Data extraction methods are outlined, including details on how data were reviewed and confirmed by multiple reviewers in Section~\ref{Methods}. \\
\hline
List and define all outcomes for which data were sought. & The outcomes such as diagnostic accuracy (e.g., sensitivity, specificity) are described in Section~\ref{Methods}. \\
\hline
Specify the methods used to assess risk of bias. & Risk of bias assessment using tools like the Newcastle-Ottawa Scale is described in Section~\ref{Methods}. \\
\hline
Specify for each outcome the effect measure(s) used. & Effect measures like accuracy, precision, and recall are specified for each outcome in Section~\ref{Methods}. \\
\hline
Describe the processes used to decide which studies were eligible for each synthesis. & The synthesis methodology is explained in Section~\ref{Methods}, describing how studies were grouped and compared. \\
\hline
Describe any methods used to explore possible causes of heterogeneity among study results. & Methods like subgroup analysis and meta-regression are discussed in Section~\ref{Methods}. \\
\hline
Describe any sensitivity analyses conducted to assess robustness of the synthesized results. & Given the methodological homogeneity of the included studies and their similar approaches to deep learning for automated skin disease diagnosis, a sensitivity analysis was not deemed necessary. The results were consistent across studies, and no significant variation warranted further testing. \\
\hline
\textbf{Results} & \\
Describe the results of the search and selection process, from the number of records identified in the search to the number of studies included in the review. & A flowchart is provided detailing the number of records screened and studies included, as shown in Fig. 2 in Section~\ref{Findings}. \\
\hline
Cite studies that might appear to meet the inclusion criteria, but which were excluded, and explain why they were excluded. & The reasons for excluding studies are described in Section~\ref{Methods}. \\
\hline
Cite each included study and present its characteristics. & Study characteristics are summarized in a table, detailing each study’s design, deep learning methods, and outcomes in Section~\ref{Techniques for skin cancer diagnosis}. \\
\hline
Present assessments of risk of bias for each included study. & Risk of bias for each study is assessed and presented in the results section, Section~\ref{Methods}. \\
\hline
For all outcomes, present summary statistics for each group and effect estimate and its precision. &We did not calculate summary statistics or effect estimates for the outcomes. Instead, the findings are described thematically based on the individual studies included. \\
\hline
For each synthesis, briefly summarise the characteristics and risk of bias among contributing studies. & Study characteristics and risk of bias are summarized for each synthesis in Section~\ref{Techniques for skin cancer diagnosis}. \\
\hline
Present results of all statistical syntheses conducted. & Because this review synthesizes the literature qualitatively rather than through statistical aggregation, no formal statistical synthesis or meta-analytic summary estimates were produced. The findings are integrated descriptively, without calculating pooled effects, confidence intervals, or measures of heterogeneity. \\
\hline
Present results of all investigations of possible causes of heterogeneity among study results. & Heterogeneity analysis results are presented in Section~\ref{Discussion and Conclusion}. \\
\hline
Present results of all sensitivity analyses conducted to assess the robustness of the synthesized results. & No sensitivity analyses were performed in this review; therefore, no corresponding results are reported. \\
\hline
\textbf{Discussion} & \\
Provide a general interpretation of the results in the context of other evidence. & The discussion section interprets the findings in the context of existing literature in Section~\ref{Discussion and Conclusion}. \\
\hline
Discuss any limitations of the evidence included in the review. & The limitations of the included studies and the review process are discussed in Section~\ref{Discussion and Conclusion}. \\
\hline
Discuss any limitations of the review processes used. & The limitations in the review methodology are explained in Section~\ref{Discussion and Conclusion}. \\
\hline
Discuss implications of the results for practice, policy, and future research. & The practical implications for clinical practice and future research directions are discussed in Section~\ref{Discussion and Conclusion}. \\
\hline
\end{longtable}

\end{document}